\PassOptionsToPackage{table}{xcolor}
\documentclass{article}
\usepackage{colm2024_conference}
\usepackage{microtype}
\usepackage{hyperref}
\usepackage{url}
\usepackage{booktabs}
\usepackage{color, colortbl}
\usepackage[table]{xcolor} 
\usepackage{amsmath}
\usepackage{amsfonts}
\usepackage{subcaption}
\usepackage{multicol}
\usepackage{multirow}
\usepackage{cleveref}
\usepackage[first=0,last=9]{lcg}

\title{LLM2Vec: Large Language Models Are Secretly Powerful Text Encoders}

\author{Parishad BehnamGhader$^{*, \diamond}$\thanks{$^{*}$Equal contribution.}~~~ Vaibhav Adlakha$^{*, \diamond, \dagger}$~~~ Marius Mosbach$^{\diamond}$\\
\medskip
\textbf{Dzmitry Bahdanau}$^{\dagger}$~~~ \textbf{Nicolas Chapados}$^{\dagger}$~~~ \textbf{Siva Reddy}$^{\diamond, \dagger, \ddagger}$\\
\medskip
$^{\diamond}$McGill University, Mila~~~ $^{\dagger}$ServiceNow Research~~~ $^{\ddagger}$Facebook CIFAR AI Chair\\
\texttt{\{parishad.behnamghader,vaibhav.adlakha,marius.mosbach\}@mila.quebec} \\
}

\usepackage{tabularray}
\usepackage{cellspace}
\usepackage{setspace}

\usepackage{todonotes}
\usepackage[normalem]{ulem}

\makeatletter
\newcommand*\iftodonotes{\if@todonotes@disabled\expandafter\@secondoftwo\else\expandafter\@firstoftwo\fi}  
\makeatother

\newcommand{\note}[4][]{\todo[author=#2,color=#3,size=\scriptsize,fancyline,caption={},#1]{#4}} 

\newcommand{\parishad}[2][]{\note[#1]{Parishad}{teal!40}{#2}}

\newcommand{\R}{\mathcal{R}}

\newcommand{\sllama}{\texttt{S-LLaMA-1.3B}}
\newcommand{\llama}{\texttt{LLaMA-2-7B}}
\newcommand{\mistral}{\texttt{Mistral-7B}}
\newcommand{\llamathree}{\texttt{Meta-LLaMA-3-8B}}

\colmfinalcopy 

\makeatletter
\def\thanks#1{\protected@xdef\@thanks{\@thanks
        \protect\footnotetext{#1}}}
\makeatother

\begin{document}

\maketitle


\begin{abstract}

Large decoder-only language models (LLMs) are the state-of-the-art models on most of today's NLP tasks and benchmarks. Yet, the community is only slowly adopting these models for text embedding tasks, which require rich contextualized representations. In this work, we introduce LLM2Vec, a simple unsupervised approach that can transform any decoder-only LLM into a strong text encoder. LLM2Vec consists of three simple steps: 1) enabling bidirectional attention, 2) masked next token prediction, and 3) unsupervised contrastive learning. We demonstrate the effectiveness of LLM2Vec by applying it to 4 popular LLMs ranging from 1.3B to 8B parameters and evaluate the transformed models on English word- and sequence-level tasks. We outperform encoder-only models by a large margin on word-level tasks and reach a new unsupervised state-of-the-art performance on the Massive Text Embeddings Benchmark (MTEB). Moreover, when combining LLM2Vec with supervised contrastive learning, we achieve state-of-the-art performance on MTEB among models that train only on publicly available data (as of May 24, 2024). Our strong empirical results and extensive analysis demonstrate that LLMs can be effectively transformed into universal text encoders in a parameter-efficient manner without the need for expensive adaptation or synthetic GPT-4 generated data.




\end{abstract}


\section{Introduction}
\label{sec:introduction}

Text embedding models aim to encode the semantic content of natural language text in vector representations which then facilitate various natural language processing (NLP) tasks, such as semantic textual similarity, information retrieval, and clustering. 
For many years, the dominating paradigm for building such models relied on pre-trained bidirectional encoders or encoder-decoders such as BERT \citep{devlin-2019-BERT}
and T5 \citep{raffel-etal-2020-t5}, which are typically adapted for text embedding tasks by following a multi-step training pipeline consisting of weakly- and fully-supervised contrastive training \citep[\textit{inter alia}]{ni-etal-2022-GTR,li-etal-2023-gte,shitao-etal-2023-bge}.
Only recently, the community started to adopt decoder-only LLMs for embedding text \citep{Muennighoff2022-SGPT, Ma2023-repllama, Wang2023-e5mistral, springer2024repetition, li-li-2024-bellm}.


We speculate that the slow adoption of decoder-only LLMs for text embedding tasks is partly due to their causal attention mechanism, which inherently limits their ability to produce rich contextualized representations.
At any given layer, causal attention limits token interactions, ensuring that the representation of a token at position $i$ is influenced solely by the representations of preceding tokens at positions $0, 1, \dots, i-1$. Although this limitation is necessary for generative capabilities, it is sub-optimal for text embeddings as it prevents the representations from capturing information across the entire input sequence.

\begin{figure}[t]
    \centering
    \begin{subfigure}[t]{0.29\textwidth}
        \raggedleft \includegraphics[width=0.95\textwidth]{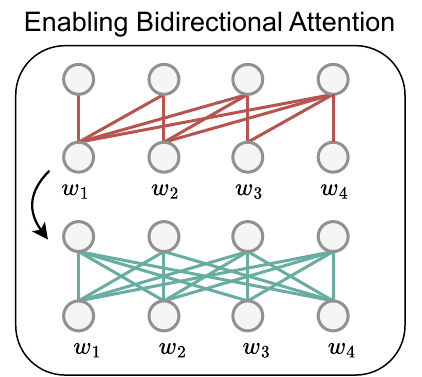}
    \end{subfigure}%
    ~~
    \begin{subfigure}[t]{0.29\textwidth}
        \centering
        \includegraphics[width=0.95\textwidth]{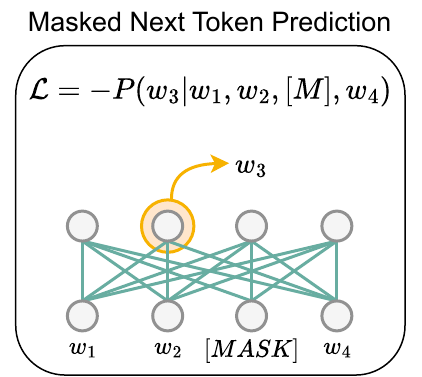}
    \end{subfigure}%
    ~~
    \begin{subfigure}[t]{0.29\textwidth}
        \raggedright
        \includegraphics[width=0.95\textwidth]{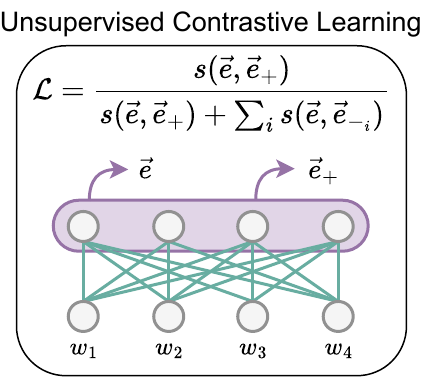}
    \end{subfigure}%
    \caption{
    The 3 steps of LLM2Vec. First, we enable bidirectional attention to overcome the restrictions of causal attention \textbf{(Bi)}. Second, we adapt the model to use bidirectional attention by masked next token prediction training \textbf{(MNTP)}. Third, we apply unsupervised contrastive learning with mean pooling to learn better sequence representations \textbf{(SimCSE)}.
    }
    \label{fig:llm2vec}
\end{figure}

Overcoming this architectural limitation of decoder-only LLMs for text embedding tasks is highly appealing as these models come with several advantages compared to their encoder-only counterparts.\footnote{We acknowledge that there are also several challenges associated with the large size of these models and provide a discussion in \Cref{sec:appendix:limitations}.} During pre-training, decoder-only LLMs learn from all input tokens and not just a small percentage\footnote{Encoder-only models are typically pre-trained by masking 15\% of the tokens in the input sequence \citep{devlin-2019-BERT}.}, which---given the same amount of training data---makes them much more sample-efficient than encoder-only models \citep{clark2020electra}. Moreover, there exists a rich ecosystem around these models, with extensive tooling and well tested pre-training recipes, which has resulted in continuous improvement of these models by the community.
Lastly, recent work on instruction fine-tuning and learning from human preferences has resulted in decoder-only LLMs that excel at instruction following \citep{wang-etal-2022-superni, ouyang2022training}, making them an ideal choice for building \textit{universal text embedding models} that generalize across a large variety of tasks using instructions.


In this work, we provide a simple unsupervised approach, termed \textbf{LLM2Vec}, which can be used to transform \emph{any} pre-trained decoder-only LLM into a (universal) text encoder. As shown in \Cref{fig:llm2vec}, LLM2Vec consists of three simple steps: 1) enabling bidirectional attention, 2) masked next token prediction, and 3) unsupervised contrastive learning. Crucially, LLM2Vec does not require any labeled data and is highly data- and parameter-efficient.


We apply LLM2vec to 4 decoder-only LLMs ranging from 1.3B to 8B parameters (\sllama{}, \llama{}, \mistral{}, \llamathree{}) and evaluate the resulting models on word- and sequence-level tasks. On word-level tasks (chunking, named-entity recognition, and part-of-speech tagging), LLM2Vec-transformed models outperform strong encoder-only models by a large margin, demonstrating its effectiveness for producing rich contextualized token representations. On the Massive Text Embeddings Benchmark (MTEB), LLM2Vec-transformed models set a new state-of-the-art for unsupervised models, with our best model reaching a score of $56.8$.
Additionally, we combine LLM2Vec with supervised contrastive training and achieve a new state-of-the-art performance among models that train only 
on publicly available data. Beyond our strong empirical results, we provide an extensive analysis of how LLM2Vec affects the representations of the underlying model and reveal an intriguing property of Mistral-7B, which can handle bidirectional attention without any fine-tuning. 

Overall, our work demonstrates that decoder-only LLMs are indeed capable of producing universal text embedding and only very little adaptation is required to reveal this ability. Our code and pre-trained models is publicly available at \url{https://github.com/McGill-NLP/llm2vec}.

\section{LLM2Vec}
\label{sec:llm2vec}
\subsection{Three simple ingredients}

\paragraph{Enabling bidirectional attention}

The first step of the LLM2Vec approach is to replace the causal attention mask of decoder-only LLMs by an all-ones matrix (see \Cref{sec:appendix:self-attention} for background on the self-attention). 
This gives each token access to every other token in the sequence, converting it into a bidirectional LLM \citep{devlin-2019-BERT, liu2019-roberta}. However, it is not a priori clear, why this should lead to better sequence representations. After all, the decoder-only LLM was not trained to attend to future tokens and therefore, this naive approach might even lead to worse representations.  As we show, simply enabling bidirectional attention does indeed decrease in embedding performance for most models. We can however easily adapt a model to make use of its bidirectional attention.

\paragraph{Masked next token prediction} 

We use a simple strategy to make the model aware of its bidirectional attention by adapting it via \textit{masked next token prediction} (MNTP). 
MNTP is a training objective that combines next token prediction with masked language modeling \citep{lv-etal-2023-falling}. Given an arbitrary sequence $\mathbf{x} = (x_1, x_2, \ldots, x_N)$ as input, we first mask a fraction of the input tokens and then train the model to predict the masked tokens based on the past and future context. Crucially, when predicting a masked token at position $i$, we compute the loss based on the logits obtained from the token representation at the previous position $i-1$, not the masked position itself (see \Cref{fig:llm2vec}). 



\paragraph{Unsupervised contrastive learning}
While the previous two steps of the LLM2Vec recipe can transform any decoder-only LLM into an encoder for word-level tasks, they might not be sufficient for sequence representations. Unlike bidirectional encoders that include a next sentence prediction objective in their pre-training objectives~\citep{devlin-2019-BERT}, decoder-only LLMs are not explicitly trained to capture the context of the entire sequence. To fill this gap, we apply unsupervised contrastive learning via SimCSE \citep{gao-etal-2021-simcse}. Specifically, given an input sentence, it is passed through the model twice with independently sampled dropout masks, resulting in two different representations for the same sentence.
The model is trained to maximize the similarity between these two representations while minimizing the similarity with representations of other sentences in the batch. Crucially, this step does not require any sentence pair data and can be applied using any collection of sentences.
We use a pooling operation on the word representations to get the sentence representation (more details in \Cref{subsecec:sequence-level-evaluation}).

\subsection{Transforming decoder-only LLMs with LLM2Vec}
\label{sec:transforming}



\paragraph{Models}

For most of our results, we experiment with 3 different decoder-only LLMs ranging from 1.3B to 7B parameters: Sheared-LLaMA-1.3B (\sllama{}, \citealp{xia2023-sheared-llama}), Llama-2-7B-chat (\llama{}, \citealp{Hugo2023-llama-2}), and Mistral-7B-Instruct-v0.2 (\mistral{}, \citealp{jiang2023-mistral}). In \Cref{tab:mteb-unsupervised,tab:supervised}, we provide additional results for the recently released Meta-Llama-3-8B-Instruct model (\llamathree{}, \citealp{llama3}).


\paragraph{Training data}


We perform both the MNTP and the unsupervised SimCSE step using data from English Wikipedia. We select data from Wikipedia as it is presumably included in the pre-training mixture of all the models we experiment with. It is therefore fair to assume that these two adaptation steps are not teaching the model any new knowledge beyond how to attend to future tokens and how to construct sequence representations. Specifically, we use the Wikitext-103 dataset \citep{merity2017pointer} for the MNTP step and a subset of Wikipedia sentences released by \cite{gao-etal-2021-simcse} for the unsupervised SimCSE step.

\paragraph{Masked next token prediction}

We follow established practice from the masked language modeling literature and randomly mask a fraction of the tokens from the input sequence \citep{devlin-2019-BERT, liu2019-roberta}. We use the underscore (\texttt{\_}) as the mask token, since the models we experiment with do not have a special token for masking. We fine-tune the model using LoRA \citep{hu2022lora} to predict the masked token using the representation of the previous token to maximally align our training objective with the pre-training setup of decoder-only LLMs. For all models, we trained for 1000 steps with a batch size of 32 on a single 80GB A100 GPU. For 7B and 8B models, this training takes only 100 minutes. We provide additional details of our training setup and hyperparameters in \Cref{subsec:mntp-training-details}.

\paragraph{Unsupervised contrastive learning}


For the contrastive training, we apply the unsupervised SimCSE approach by \citet{gao-etal-2021-simcse}. The positive examples are constructed by applying LLM's dropout twice on the same input sequence, whereas the other sequences in the batch act as in-batch negatives. We merge the MNTP LoRA weights into the base model and initialize new LoRA parameters before starting the SimCSE training, which ensures that the models retains the knowledge learned in the previous step.
Similar to the MNTP step, we train for 1000 steps.
For 7B and 8B models, this training takes 3 hours on a single 80GB A100 GPU with a batch size of 128. We provide additional details of our training setup and hyperparameters in \Cref{subsec:simcse-training-details}.




\section{LLM2Vec-transformed models are strong unsupervised text embedders}
\label{sec:unsupervised}


\subsection{Evaluation on word-level tasks}
\label{subsecec:word-level-evaluation}

We start by evaluating on word-level tasks to demonstrate that LLM2Vec is successful at improving the contextual representations constructed by decoder-only LLMs.


\begin{figure}[t]
    \centering
    \begin{subfigure}[t]{0.33\textwidth}
        \centering
        \includegraphics[width=0.90\textwidth]{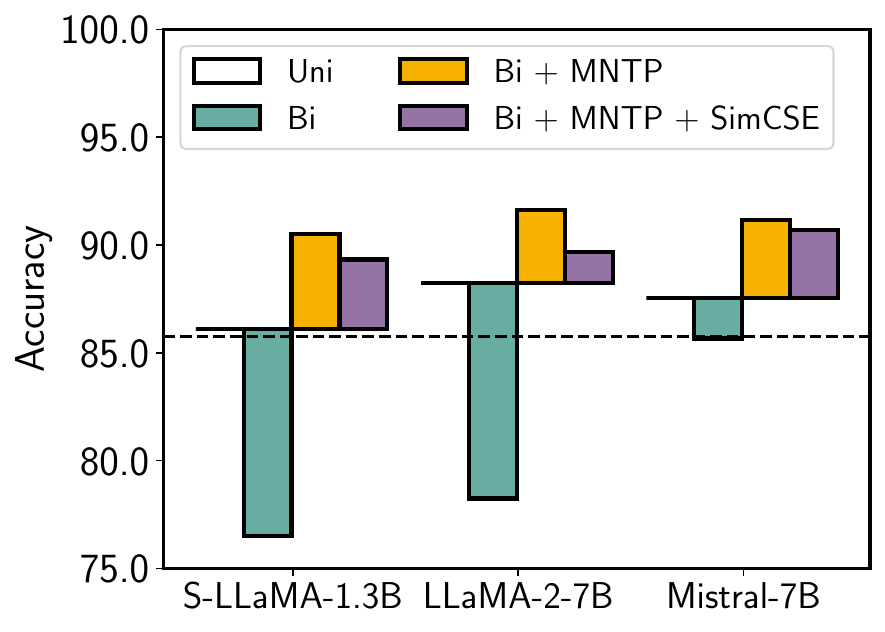}
        \caption{Chunking}
    \end{subfigure}%
    ~
    \begin{subfigure}[t]{0.33\textwidth}
        \centering
        \includegraphics[width=0.90\textwidth]{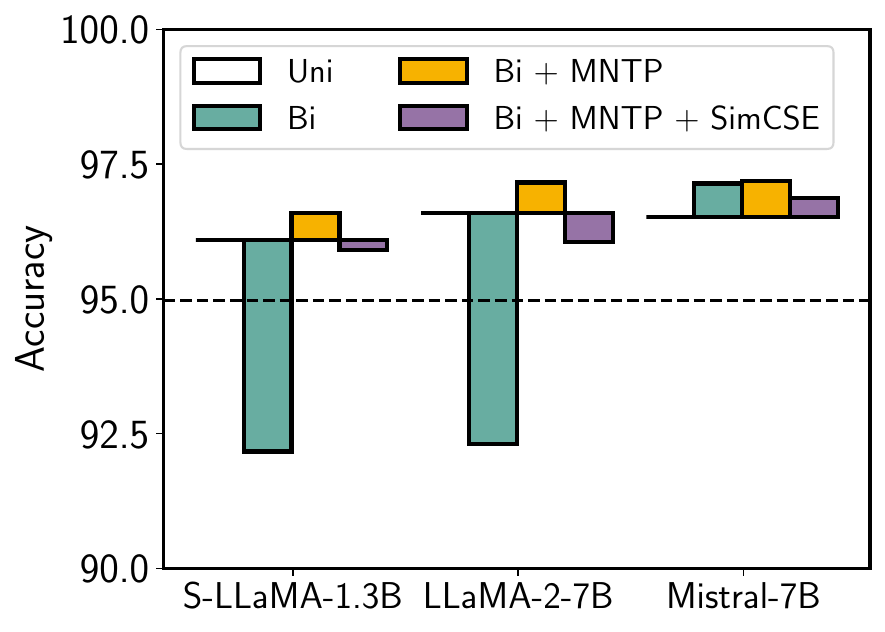}
        \caption{NER}
    \end{subfigure}%
    ~
    \begin{subfigure}[t]{0.33\textwidth}
        \centering
        \includegraphics[width=0.90\textwidth]{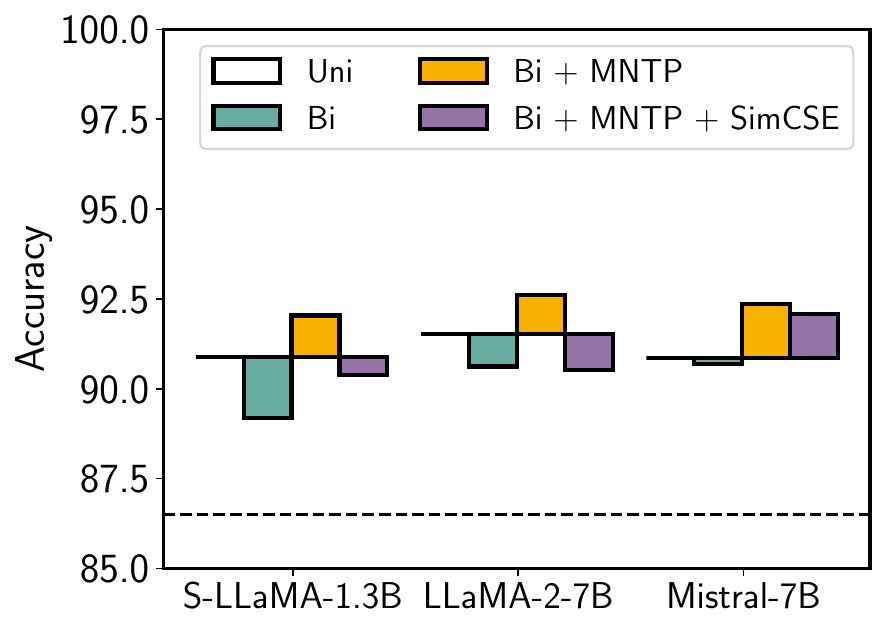}
        \caption{POS}
    \end{subfigure}    
    \caption{Evaluation of LLM2Vec-transformed models on word-level tasks. Solid and dashed horizontal lines show the performance of \texttt{Uni} and \texttt{DeBERTa-v3-large}, respectively.
    }
    \label{fig:word-level-results}
\end{figure}

\paragraph{Setup}

We evaluate three word-level tasks: chunking, named-entity recognition (NER), and part-of-speech tagging (POS), using the CoNLL-2003 benchmark~\citep{conll-2003}. We embed each input sentence and train a task-specific linear classifier on top of the frozen representations. This is akin to the linear probing setup commonly used in the language model analysis literature \citep{belinkov-2022-probing}.
We compare the LLM2Vec-transformed models to \texttt{DeBERTa-v3-large} \citep{he2023debertav}, the current state-of-the-art encoder-only model. Additional details about our setup are provided in \Cref{subsec:word-level-training-details}. 

\paragraph{Results}


\Cref{fig:word-level-results} shows the results of our evaluation (a detailed breakdown of the results is provided in \Cref{tab:appendix:word-task-results}).
On each of the three tasks, constructing token representations with causal attention (Uni) already outperforms the encoder-only baseline. This is not surprising, given that the models we experiment with are significantly larger and have been pre-trained on more data. As expected, naively applying bidirectional attention dramatically hurts performance in most cases. Interestingly, for \mistral{}, enabling bidirectional attention hurts performance much less compared to \sllama{} and \llama{}. For NER, Mistral's performance even improves by $0.6\%$ with bidirectional connections.



Focusing on the LLM2Vec-transformed models, we observe that for all models and tasks, adapting via MNTP improves performance. For instance, in the chunking task, we see improvements for \sllama{} (by $5\%$), \llama{} (by $4\%$), and \mistral{} (by $4\%$). Combining MNTP with SimCSE, however, performs worse than just applying MNTP. This is expected for word-level tasks, as SimCSE adapts the representations for sequence-level tasks.\looseness-1 

\subsection{Evaluation on sequence-level tasks}
\label{subsecec:sequence-level-evaluation}


\begin{figure}[t]
    \centering
    \begin{subfigure}[t]{0.33\textwidth}
        \centering
        \includegraphics[width=0.90\textwidth]{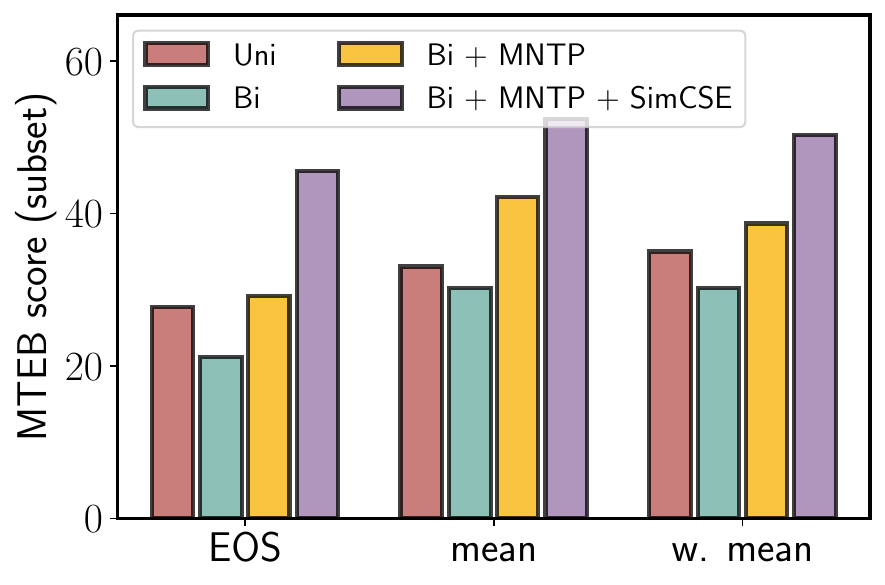}
        \caption{\texttt{S-LLaMA-1.3B}}
    \end{subfigure}%
    ~~
    \begin{subfigure}[t]{0.33\textwidth}
        \centering
        \includegraphics[width=0.90\textwidth]{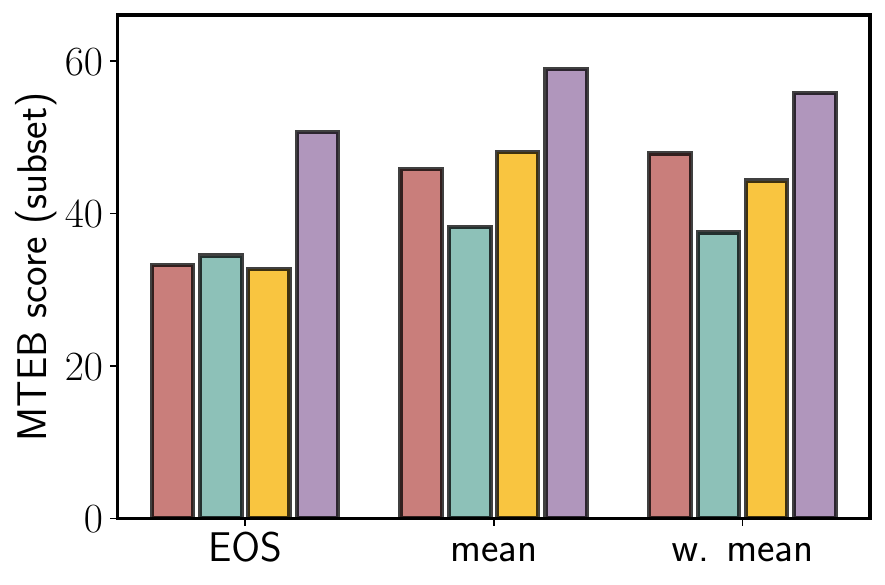}
        \caption{\texttt{Llama-2-7B}}
    \end{subfigure}%
    ~~
    \begin{subfigure}[t]{0.33\textwidth}
        \centering
        \includegraphics[width=0.90\textwidth]{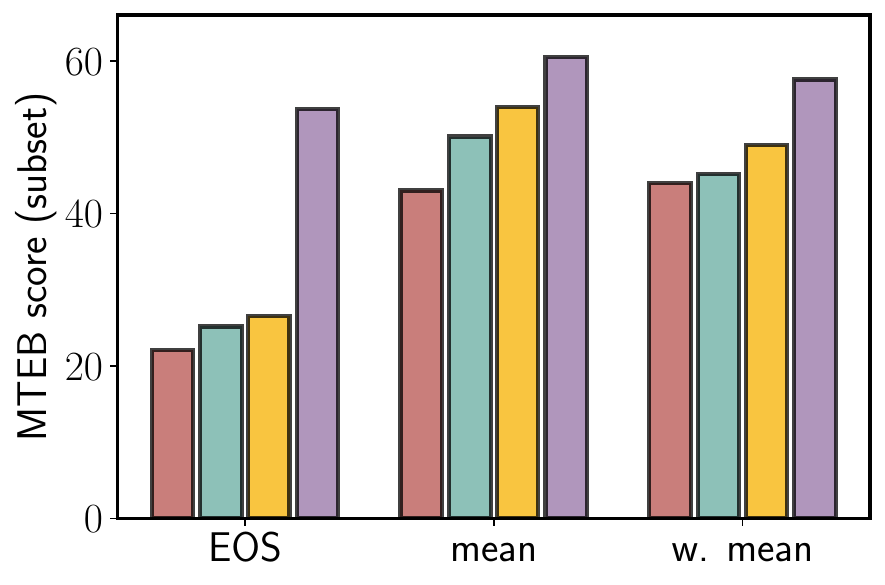}
        \caption{\texttt{Mistral-7B}}
    \end{subfigure}    
    \caption{Unsupervised results on our 15 task subset of the MTEB dataset. We ablate three different pooling choices: EOS, mean pooling, and weighted mean pooling. LLM2Vec is compatible with all three approaches and works best with mean pooling.}
    \label{fig:mteb-subset-unsupervised}
\end{figure}


Next, we evaluate on the Massive Text Embedding Benchmark (MTEB), a collection of 7 diverse embedding task categories covering a total of 56 datasets \citep{muennighoff-etal-2023-mteb}. To select the best-performing pooling method for each method, we perform ablations on a 15 task subset consisting of representative tasks from each of the MTEB categories. We provide additional details and justification for how we chose this subset in \Cref{subsec:appendix:mteb}.

\paragraph{Setup}

Following previous work \citep{su-etal-2023-Instructor, Wang2023-e5mistral, springer2024repetition}, we evaluate with task-specific instructions.
For a fair comparison, we use the same set of instructions as \citet{Wang2023-e5mistral} which are also used by \citet{springer2024repetition}. The instructions are only added to queries and can be found in \Cref{tab:appendix:mteb_instructions} of \Cref{subsec:appendix:mteb-instruction}.
For symmetric tasks, the same instruction will be used for the query and the document. When applying (weighted) mean pooling \citep{Muennighoff2022-SGPT}, we exclude the instruction tokens. 

As a baseline, we compare to the unsupervised BERT models obtained from \citet{gao-etal-2021-simcse}. Additionally, we compare to Echo embeddings, a concurrent approach by \citet{springer2024repetition}, which we run with the same models and instructions (see \Cref{subsec:appendix:echo-reproducibility} for more details on our implementation of Echo embeddings). 
Echo duplicates the input and takes the pooling over the second occurrence to address the limitation of causal information flow.

\paragraph{Results on our 15 task subset of MTEB}


\Cref{fig:mteb-subset-unsupervised} shows the impact of various pooling methods for all three models on the subset of MTEB tasks. 
We can clearly observe that applying causal attention is sub-optimal when constructing text embeddings. The dominant paradigm of applying the EOS pooling for models with causal attention is outperformed by (weighted) mean pooling. Enabling bidirectional attention without any training harms performance for \sllama{} and \llama{}. Similar to our word-level results, the performance of \mistral{} improves with bidirectional attention, even without any training. 

For LLM2Vec-transformed models, applying MNTP training improves the performance of \sllama{} and \mistral{}. Moreover, applying SimCSE further boosts the performance of \sllama{}, \llama{}, and \mistral{} by $49.8\%$, $23.2\%$, and $37.5\%$ compared to the best causal baseline on the MTEB subset.
We further conduct an ablation of each component of LLM2Vec recipe in
\Cref{subsec:sentence-level-results} (\Cref{tab:appendix:unsupervised-pooling-results}).

\begin{table*}[t]
    \centering
    \small
    \resizebox{\textwidth}{!}{
    \begin{tabular}{lrrrrrrrr}
    \toprule
    \textbf{Categories}\phantom{...} \textbf{$\rightarrow$} & \textbf{Retr.} & \textbf{Rerank.} & \textbf{Clust.} & \textbf{PairClass.} & \textbf{Class.} & \textbf{STS} & \textbf{Summ.} & \textbf{Avg} \\
    \textbf{\# of datasets} \textbf{$\rightarrow$} & \multicolumn{1}{c}{15}     & \multicolumn{1}{c}{4}     & \multicolumn{1}{c}{11}   & \multicolumn{1}{c}{3}      & \multicolumn{1}{c}{12}    & \multicolumn{1}{c}{10}  & \multicolumn{1}{c}{1}    & \multicolumn{1}{c}{56} \\ 
    \midrule
    \multicolumn{9}{c}{\texttt{Encoder-only}}\\\midrule
    BERT & 10.59 & 43.44 & 30.12 & 56.33 & 61.66 & 54.36 & 29.82 & 38.33\\
    \multicolumn{1}{l}{BERT + SimCSE}   &  20.29  &  46.47 & 29.04  & 70.33  & 62.50 & 74.33 & 31.15  & 45.45 \\  
    \midrule
    
    \multicolumn{9}{c}{\sllama{}}\\\midrule
    Uni + w. Mean & 9.47 & 38.02 & 28.02 & 42.19 & 59.79 & 49.15 & 24.98 & 35.05 \\
    \rowcolor{lightgray!50}LLM2Vec (w/o SimCSE) &  15.48 & 40.99 & 31.83 & 50.63 & 64.54 & 62.06 & 26.82 & 41.43  \\ 
    \rowcolor{lightgray!50}LLM2Vec &  25.93 & 47.70 & 37.45 & 72.21 & 67.67 & 71.61 & 31.23 & 49.42\\
    Echo & 10.36 & 41.52 & 30.03 & 52.08 & 63.75 & 59.36 & 22.79 & 39.10\\ 
    \midrule
    
    \multicolumn{9}{c}{\llama{}}\\\midrule
    Ui + w. Mean & 15.16 & 46.94 & 36.85 & 61.41 & 69.05 & 63.42 & 26.64 & 44.54 \\
    \rowcolor{lightgray!50}LLM2Vec (w/o SimCSE) &  19.86 & 44.74 & 35.31 & 61.60 & 67.94 & 66.74 & 26.83 & 45.70  \\ 
    \rowcolor{lightgray!50}LLM2Vec & 36.75 & 52.95 & 40.83 & 77.89 & 71.57 & 76.41 & 31.38 & 55.36 \\
    Echo &  16.16 & 46.84 & 34.25 & 63.54 & 69.82 & 67.95 & 25.57 & 45.36 \\ 
    \midrule
    \multicolumn{9}{c}{\mistral{}}\\\midrule
    Uni + w. Mean & 10.43 & 45.11 & 35.82 & 60.28 & 71.14 & 58.59 & 26.57 & 42.46  \\
    Bi + Mean & 15.84 & 47.40 & 35.55 & 66.53 & 72.18 & 71.04 & 29.93 & 46.86 \\
    \rowcolor{lightgray!50}LLM2Vec (w/o SimCSE) & 19.74 & 50.43 & 40.06 & 70.95 & 72.51 & 71.90 & 27.84 & 49.43 \\ 
    \rowcolor{lightgray!50}LLM2Vec &  38.05 & \textbf{53.99} & 40.63 & \textbf{80.94} & \textbf{74.07} & \textbf{78.50} & 30.19 & \textbf{56.80}  \\ 
    Echo & 22.68 & 51.07 & 36.78 & 75.87 & 72.69 & 73.60 & 29.54 & 50.26 \\ 
    \midrule
    \multicolumn{9}{c}{\llamathree{}}\\\midrule
    Uni + w. Mean & 15.17 & 46.22 & 36.84 & 60.94 & 67.41 & 62.80 & 25.51 & 43.98 \\
    Bi + Mean & 3.90 & 34.56 & 14.27 & 42.71 & 57.89 & 51.15 & 23.26 & 30.56 \\
    \rowcolor{lightgray!50}LLM2Vec (w/o SimCSE) & 24.75 & 49.20 & 39.74 & 65.91 & 69.00 & 67.85 & 25.59 & 48.84 \\
    \rowcolor{lightgray!50}LLM2Vec & \textbf{39.19} & 53.09 & \textbf{41.99} & 78.01 & 71.88 & 75.86 & \textbf{31.45} & 56.23 \\
    Echo & 12.58 & 49.79 & 36.32 & 68.95 & 70.22 & 67.43 & 26.44 & 45.32 \\
    \bottomrule
    
    \end{tabular}
    }
    \caption{Unsupervised results on MTEB. We compare \sllama{}, \llama{}, \mistral{}, and \llamathree{} with and without LLM2Vec to the unsupervised BERT models of \citet{gao-etal-2021-simcse} as well as Echo embeddings \citep{springer2024repetition}.
    }
    \label{tab:mteb-unsupervised}
\end{table*}


\paragraph{Results on full MTEB}




\Cref{tab:mteb-unsupervised} shows the results of the best performing models, which we select based on the ablation above, on the full MTEB dataset. 
After the first two steps of LLM2Vec---bidirectional attention and MNTP---we observe a considerable improvement in performance for all four models (e.g., $16.4\%$ improvement for \mistral{}). 

When comparing to Echo embeddings, LLM2Vec (the first two steps only)\footnote{We only directly compare the performance after the first two steps of LLM2Vec to Echo embeddings as applying SimCSE involves learning sequence representation, which makes the comparison unfair.} leads to improved performance for \sllama{}, \llama{}, and \llamathree{}, and performs almost on par for \mistral{}. However, compared to Echo embeddings, LLM2Vec is much more efficient as Echo embeddings repeat the input and therefore double the sequence length which makes inference considerably slower (we provide a runtime comparison in \Cref{subsec:appendix:echo-efficiency}). Adding the final step of the LLM2Vec recipe---unsupervised SimCSE---further boosts all three models by a large margin, making our LLM2Vec \mistral{} SOTA among all unsupervised models with a score of $56.80$.


Interestingly, \llamathree{} with LLM2Vec (w/o SimCSE) outperforms echo embeddings by a larger margin compared to the other models. Adding SimCSE again boosts performance, but does not outperform LLM2Vec applied to \mistral{}. 

Overall, our results highlight that LLM2Vec is successful at transforming decoder-only LLMs into strong text embedding models which outperform previous unsupervised approaches on the challenging MTEB leaderboard.

\section{How does LLM2Vec affect a model?}
\label{sec:llm2vec-analysis}


\subsection{LLM2Vec helps models to capture information from future tokens}
\label{subsec:analysis:future-tokens}

To analyze the extent to which LLM2Vec-transformed models incorporate information from future tokens, we adopt the analysis of \citet{springer2024repetition} and test how well the model performs at judging the similarity between sentences that share the same prefix. 

\paragraph{Setup}

We evaluate on a synthetic dataset collected by \citet{springer2024repetition}, which consists of 35 sentence triples $\{(q_i$, $s^{+}_i$, $s^{-}_i)\}_{i=1}^{35}$ with $q_i = (A_i, B_i)$, $s^{+}_i = (A_i, C_i)$, and $s^{-}_i = (A_i, D_i)$, where $B_i$ and $C_i$ have a similar meaning but $B_i$ and $D_i$ don't. We compute a sequence representation for each of these sentences by pooling only over the first part of the sentence, i.e., $A_i$. We then compute the cosine similarity between the resulting embeddings. A model that incorporates information from future tokens ($B_i$, $C_i$, or $D_i$) in the representations of the prefix $A_i$ should assign a higher similarity to the positive example. 

\paragraph{Results}


\begin{figure}[t]
    \centering
    \begin{subfigure}[t]{0.49\textwidth}
        \begin{subfigure}[t]{0.49\textwidth}
            \centering
            \includegraphics[width=0.99\textwidth]{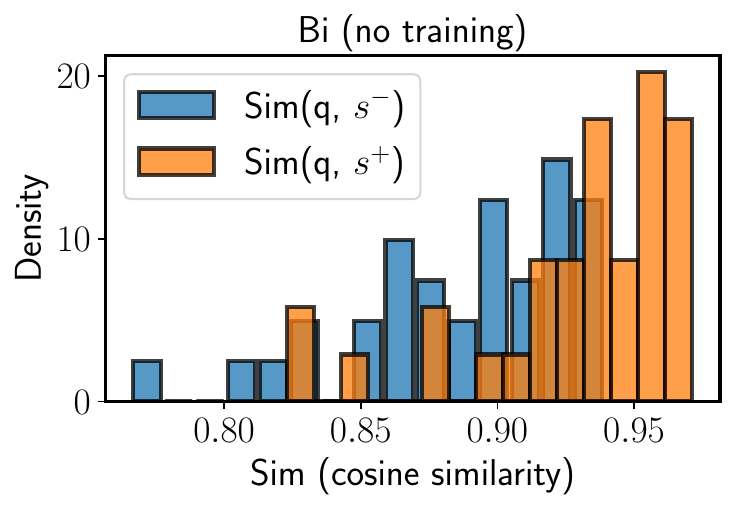}
        \end{subfigure}%
        \begin{subfigure}[t]{0.49\textwidth}
            \centering
            \includegraphics[width=0.99\textwidth]{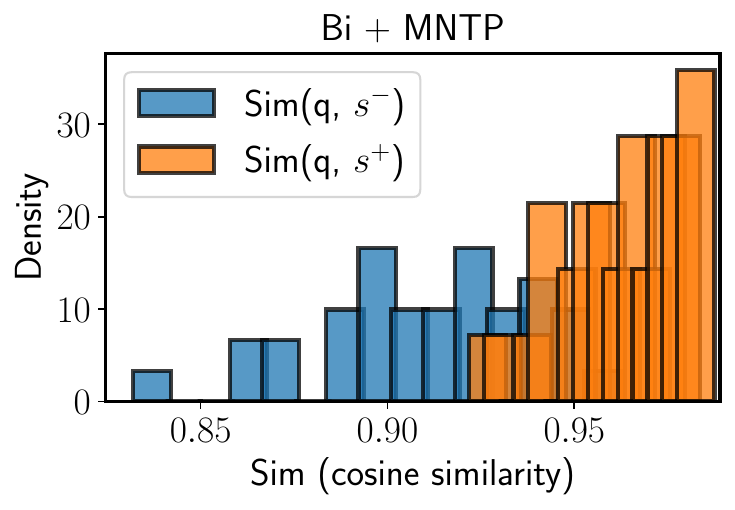}
        \end{subfigure}%
        \caption{\texttt{S-LLaMA-1.3B}}
    \end{subfigure}
    \begin{subfigure}[t]{0.49\textwidth}
        \begin{subfigure}[t]{0.49\textwidth}
            \centering
            \includegraphics[width=0.99\textwidth]{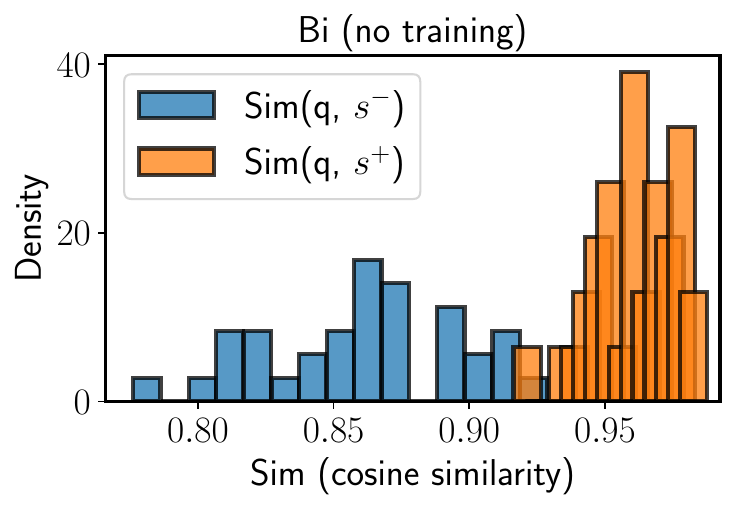}
        \end{subfigure}%
        \begin{subfigure}[t]{0.49\textwidth}
            \centering
            \includegraphics[width=0.99\textwidth]{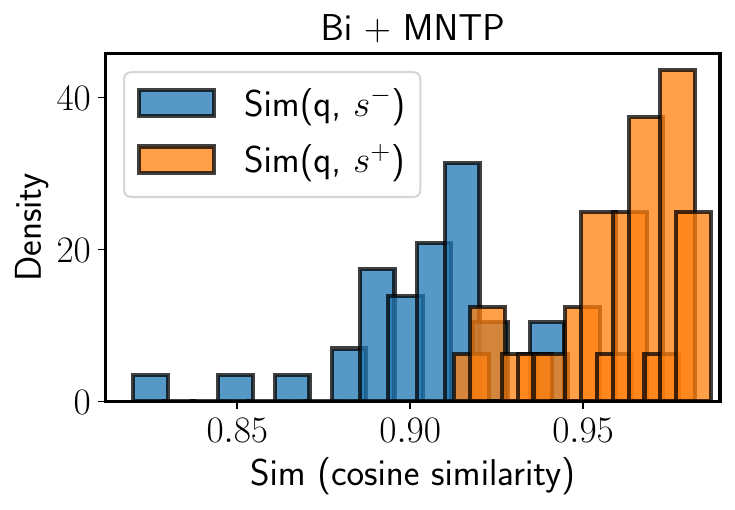}
        \end{subfigure} 
        \caption{\texttt{Mistral-7B}}
    \end{subfigure}
    \caption{Cosine similarity between query ($q$) and negative ($s^-$) as well as positive examples ($s^+$).
    Plots for \llama{} and other approaches are shown in \Cref{sec:appendix:analysis}.
    }
    \label{fig:cosine-analysis}
\end{figure}

\begin{figure}[t]
    \vspace{-1em}
    \centering
    \begin{subfigure}[t]{0.33\textwidth}
        \centering
        \includegraphics[width=0.90\textwidth]{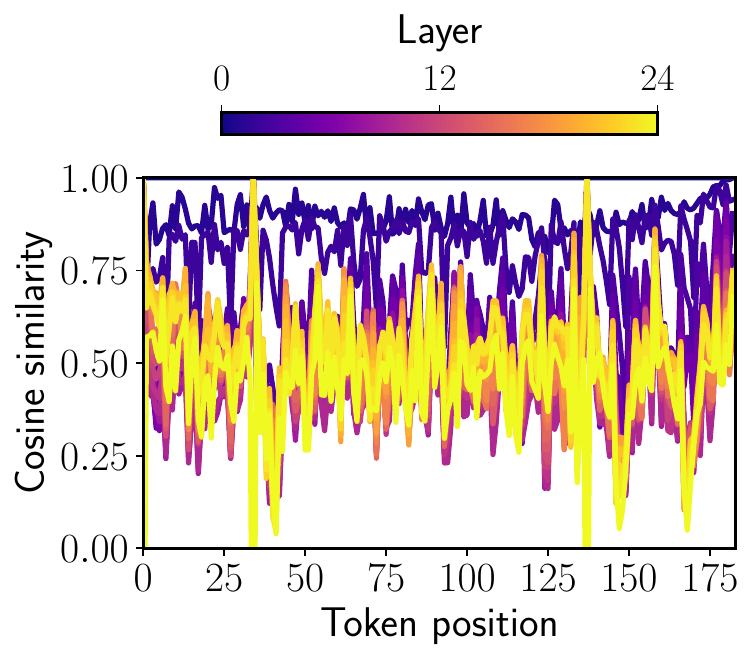}
        \caption{\texttt{S-LLaMA-1.3B}}
    \end{subfigure}%
    ~
    \begin{subfigure}[t]{0.33\textwidth}
        \centering
        \includegraphics[width=0.90\textwidth]{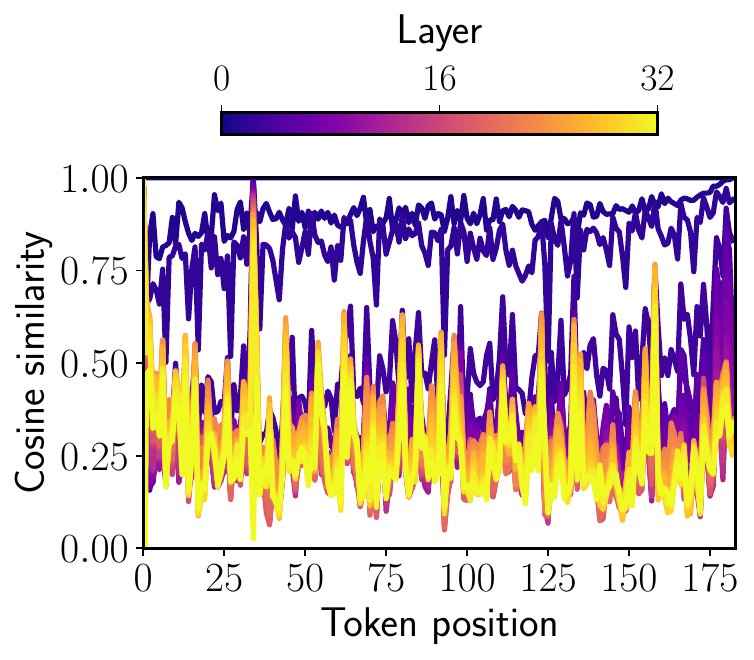}
        \caption{\texttt{Llama-2-7B}}
    \end{subfigure}%
    ~
    \begin{subfigure}[t]{0.33\textwidth}
        \centering
        \includegraphics[width=0.90\textwidth]{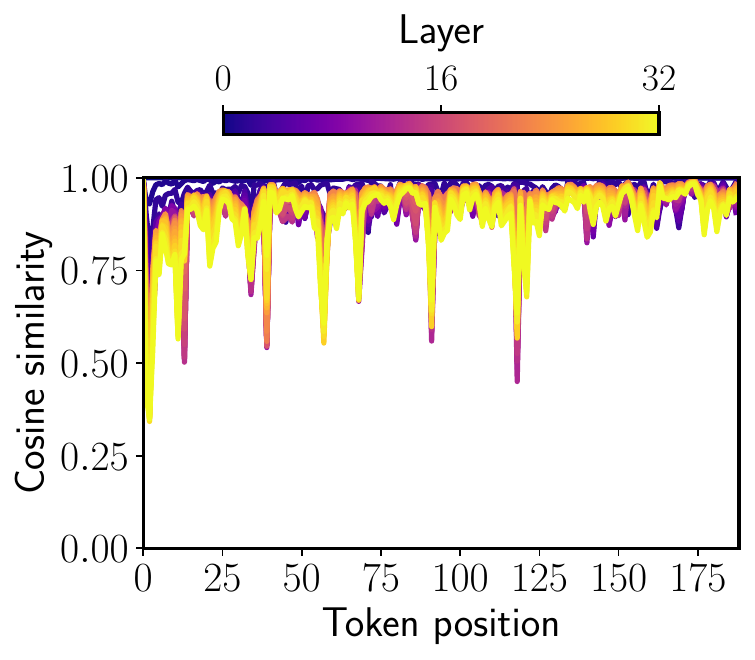}
        \caption{\texttt{Mistral-7B}}
    \end{subfigure}    
    \caption{Cosine similarities at different token positions at layers when comparing representations constructed with causal attention to those constructed with bidirectional attention (without training). Additional plots are shown in \Cref{sec:appendix:analysis}.
    }
    \label{fig:hidden_analysis}
    \vspace{-1em}
\end{figure}


\Cref{fig:cosine-analysis} shows the results of our analysis for \texttt{S-LLaMA-1.3B} and \texttt{Mistral-7B}. Results for \texttt{LLaMA-2-7B}, which show the same trends, and a comparison to Echo embeddings are provided in \Cref{sec:appendix:analysis}. For \texttt{S-LLaMA-1.3B}, we observe that enabling bidirectional attention and training with the MNTP objective are sufficient to establish a clear separation between the positive and negative examples. 
For \texttt{Mistral-7B}, all setups lead to a larger cosine similarity between the query and positive than the query and negative examples. 

\subsection{Why does bidirectional attention without training work for Mistral models?}
\label{subsec:analysis:mistral}


Our empirical results so far as well as the analysis above share an intriguing observation: enabling bidirectional attention works well for \mistral{}, even without any training. Below, we investigate this surprising behavior by analyzing how bidirectional attention impacts the representations of a model. 

\paragraph{Setup}

We feed a single input sequence (a random paragraph from Wikipedia) to each model and compute the hidden representations of every token at every layer $l$ with causal ($\mathbf{H}^c_{l}$) and bidirectional attention ($\mathbf{H}^{bi}_{l}$). For every layer, we compute the cosine similarity between the representations constructed using causal and bidirectional attention, i.e., $\text{sim}(\mathbf{H}^c_{l}, \mathbf{H}^{bi}_{l})$. For most layers, we expect this similarity to be low, as enabling bidirectional attention without any training should lead to substantially different representations.

\paragraph{Results}


\Cref{fig:hidden_analysis} shows that as expected, for \sllama{} and \llama{}, enabling bidirectional attention without training has a profound impact on the representations, leading to low cosine similarity across almost all layers and token positions. For \mistral{}, on the other hand, the representations have very high cosine similarity throughout. 

Based on these findings (we replicate these results for other inputs and other variants of Mistral in \Cref{sec:appendix:analysis}) and the strong unsupervised results for \mistral{} with bidirectional attention, we speculate that Mistral models are pre-trained with some form bidirectional attention, e.g., prefix language modeling \citep{raffel-etal-2020-t5} -- at least for some parts of its training. We leave a more detailed investigation of this intriguing behavior for future work.

\section{Combining LLM2Vec with supervised contrastive learning}
\label{sec:llm2vec+contrastive}

The final piece of our evaluation combines LLM2Vec with supervised contrastive learning.  

\subsection{LLM2Vec leads to strong performance on the MTEB leaderboard}


\paragraph{Setup}

For supervised training, we train on a replication of the public portion of the E5 dataset \citep{Wang2023-e5mistral} curated by \cite{springer2024repetition}. The dataset consists of approximately 1.5M samples and we provide details on its compilation in \Cref{sec:appendix:e5}. We follow standard practice and train the models with contrastive learning
using hard negatives and in-batch negatives.
We use LoRA fine-tuning for supervised setting as well. The MNTP LoRA weights are merged into the base model, and the trainable LoRA weights are initialized with SimCSE weights.
For LLM2Vec models that use just MNTP, the LoRA weights are randomly initialized.
The training is performed for 1000 steps with a batch size of 512.
We detail other hyperparameters in \Cref{subsec:sup-training-details}.



For a fair comparison, we only compare to models trained on publicly available data and provide a comparison to the top entries on the MTEB leaderboard in \Cref{sec:appendix:sup-mteb-results}. 

\paragraph{Results}

\begin{table*}[t]
    \centering
    \small
    \resizebox{\textwidth}{!}{
    \begin{tabular}{lcccccccc}
    \toprule
    \textbf{Categories}\phantom{...} \textbf{$\rightarrow$} & \textbf{Retr}. & \textbf{Rerank.} & \textbf{Clust.} & \textbf{PairClass.} & \textbf{Class.} & \textbf{STS} & \textbf{Summ.} & \textbf{Avg} \\
    \textbf{\# of datasets} $\rightarrow$ & \multicolumn{1}{c}{15}     & \multicolumn{1}{c}{4}     & \multicolumn{1}{c}{11}   & \multicolumn{1}{c}{3}      & \multicolumn{1}{c}{12}    & \multicolumn{1}{c}{10}  & \multicolumn{1}{c}{1}    & \multicolumn{1}{c}{56}  \\ 
    \midrule
    \multicolumn{9}{c}{\texttt{Previous work w/ public data only}}\\\midrule
    \multicolumn{1}{l}{Instructor-xl} & 49.26 & 57.29 & 44.74 & 86.62 & 73.12 & 83.06 & \textbf{32.32} & 61.79\\
    \multicolumn{1}{l}{BGE$_\text{large-en-v1.5}$}   & 54.29 & 60.03 & 46.08 & 87.12 & 75.97 & 83.11 & 31.61 & 64.23 \\ 
    GritLM$_\text{Mistral-7b-v1}$ + public data & 53.10 & \textbf{61.30} & \textbf{48.90} & 86.90 & 77.00 & 82.80 & 29.40 & 64.70 \\
    \multicolumn{1}{l}{E5$_\text{Mistral-7b-v1}$ + public data}  & 52.78 & 60.38 & 47.78 & \textbf{88.47} & 76.80 & 83.77 & 31.90 & 64.56 \\
    Echo$_\text{Mistral-7b-v1}$ & 55.52 & 58.14 & 46.32 & 87.34 & \textbf{77.43} & 82.56 & 30.73 & 64.68\\
    \midrule

    \multicolumn{9}{c}{\sllama{}}\\\midrule
    Uni + w. Mean &  51.02 & 54.65 & 39.90 & 83.57 & 71.64 & 82.16 & 30.05 & 60.44 \\
    \rowcolor{lightgray!50}LLM2Vec (w/o SimCSE) & 51.44 & 55.38 & 43.57 & 86.20 & 72.21 & 83.58 & 30.01 & 61.85 \\
    \rowcolor{lightgray!50}LLM2Vec &  51.49 & 55.58 & 43.24 & 85.80 & 72.98 & 83.62 & 30.12 & 61.96\\
    \midrule
    \multicolumn{9}{c}{\llama{}}\\\midrule
    Uni + w. Mean & 54.33 & 58.01 & 40.57 & 87.01 & 75.60 & 83.47 & 29.68 & 62.96\\
    \rowcolor{lightgray!50}LLM2Vec (w/o SimCSE) & 54.60 & 57.38 & 45.24 & 88.03 & 76.33 & 83.73 & 28.49 & 64.14\\
    \rowcolor{lightgray!50}LLM2Vec &  54.34 & 57.70 & 45.04 & 87.87 & 76.53 & 83.43 & 28.82 & 64.04 \\
    \midrule
    \multicolumn{9}{c}{\mistral{}}\\\midrule
    Uni + w. Mean & 54.81 & 57.37 & 41.07 & 86.05 & 76.01 & 83.44 & 30.74 & 63.20 \\
    \rowcolor{lightgray!50}LLM2Vec (w/o SimCSE) & 55.99 & 58.42 & 45.54 & 87.99 & 76.63 & \textbf{84.09} & 29.96 & \textbf{64.80}\\
    \rowcolor{lightgray!50}LLM2Vec & 56.05 & 58.59 & 45.12 & 88.18 & 76.72 & 83.69 & 30.66 & 64.72 \\
    \midrule
    \multicolumn{9}{c}{\llamathree{}}\\\midrule
    Uni + w. Mean & 55.42 & 58.60 & 43.19 & 86.29 & 75.56 & 83.95 & 30.59 & 63.87 \\
    \rowcolor{lightgray!50}LLM2Vec (w/o SimCSE) & 56.63 & 59.68 & 46.45 & 87.80 & 75.92 & 83.58 & 30.94 & \textbf{65.01} \\
    \rowcolor{lightgray!50}LLM2Vec & \textbf{56.71} & 59.02 & 45.86 & 87.95 & 76.67 & 82.98 & 29.67 & 64.90 \\
    \bottomrule
     
    \end{tabular}%
    }
    \caption{Supervised results on full MTEB benchmark. 
    The best performing LLM2Vec model \texttt{Meta-LLaMA-3-8B} + LLM2Vec (w/o SimCSE) achieves a new SOTA performance among models trained only on publicly available data.
    }
    \label{tab:supervised}
\end{table*}

\begin{figure}[t]
    \vspace{-0.5em}
    \centering
    \begin{subfigure}[t]{0.33\textwidth}
        \centering
        \includegraphics[width=0.90\textwidth]{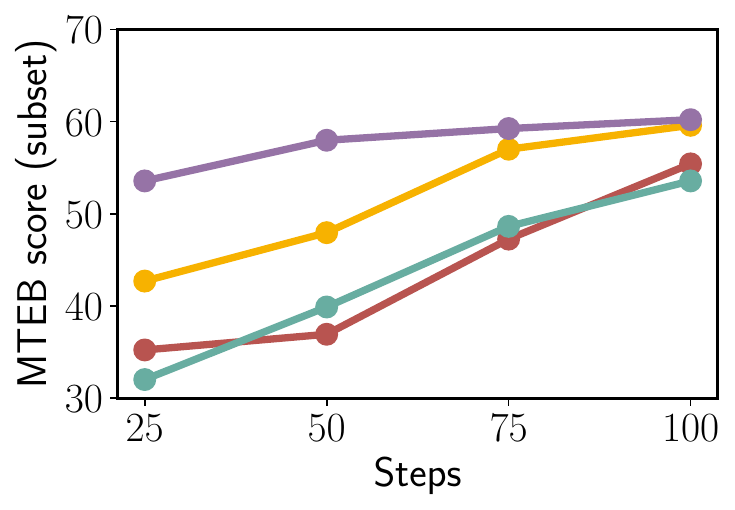}
        \caption{\texttt{S-LLaMA-1.3B}}
    \end{subfigure}%
    ~
    \begin{subfigure}[t]{0.33\textwidth}
        \centering
        \includegraphics[width=0.90\textwidth]{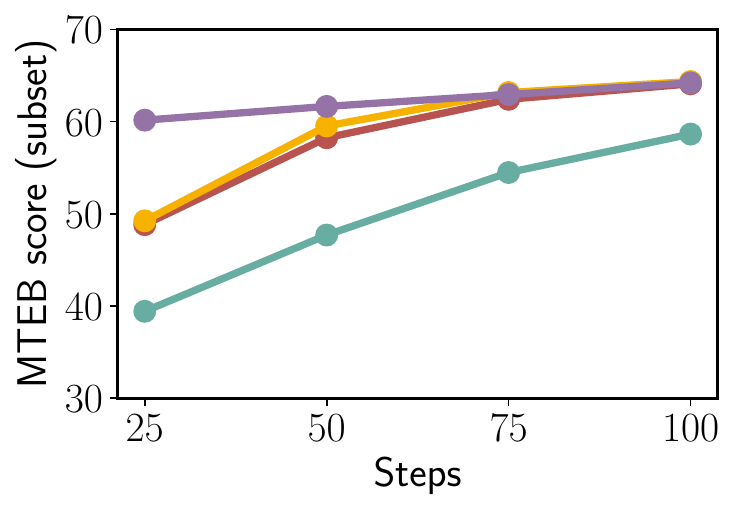}
        \caption{\texttt{Llama-2-7B}}
    \end{subfigure}%
    ~
    \begin{subfigure}[t]{0.33\textwidth}
        \centering
        \includegraphics[width=0.90\textwidth]{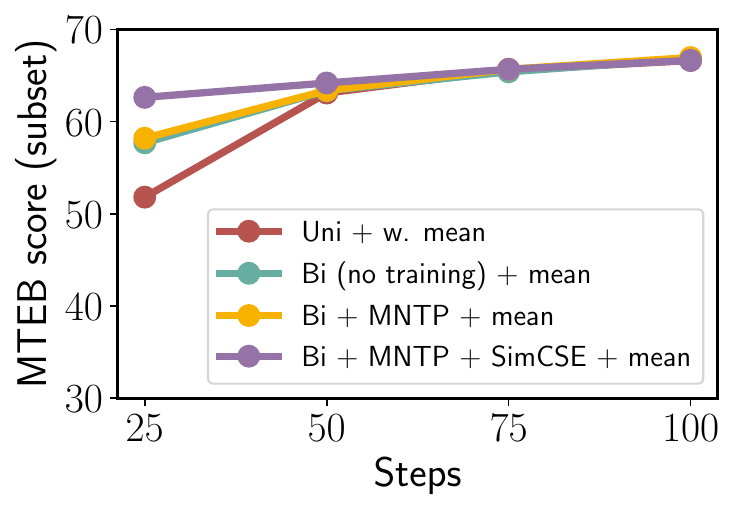}
        \caption{\texttt{Mistral-7B}}
    \end{subfigure}    
    \caption{Results on the 15 task subset of MTEB during training of \sllama{}, \llama{}, and \mistral{}. For all three models, applying LLM2Vec before supervised training leads to better performance with less steps.
    }
    \label{fig:supervised-sample-efficiency}
    \vspace{-0.5em}
\end{figure}

\Cref{tab:supervised} shows the results of our evaluation. For all models, transforming a model with LLM2Vec leads to improved performance over the strong \texttt{Uni} + weighted mean baseline. As expected, performing unsupervised SimCSE is less crucial for supervised training, and even leads to slightly worse performance for \llama{}, \mistral{}, and \llamathree{} compared to just performing the MNTP step of LLM2Vec (LLM2Vec w/o SimCSE). However, as we will show in \Cref{subsec:sample-efficiency}, LLM2Vec with MNTP and SimCSE is much more sample-efficient, and therefore crucial in compute or data-constrained settings. Notably, our best model, \llamathree{} + LLM2Vec (w/o SimCSE) leads to a new state-of-the-art performance among models trained only on publicly available data.

\subsection{LLM2Vec leads to more sample-efficient training}
\label{subsec:sample-efficiency}

\paragraph{Setup}

To demonstrate the sample-efficiency of LLM2Vec-transformed models, we save a checkpoint every 25 training steps and evaluate them on our 15 task subset of MTEB. 

\paragraph{Results}

As shown in \Cref{fig:supervised-sample-efficiency}, LLM2Vec-transformed models reach better performance earlier in training. This observation is consistent across all three models. For \sllama{}, the smallest of our three models, even performing just MNTP leads to a considerably improved sample-efficiency. These results are particularly encouraging for settings where it is hard to acquire high quality labeled data, a setting which we leave for future work.

\section{Related Work}
\label{sec:related-work}





\paragraph{Supervised text encoders}

Initially, supervised methods primarily relied on tasks such as natural language inference or sentence similarity to train BERT-like models for producing sentence embeddings \citep{conneau-etal-2017-supervised,reimers-gurevych-2019-sbert}.
Subsequently, BERT-like models have also been adapted to tasks like retrieval \citep{karpukhin-etal-2020-DPR,khattab-2020-ColBERT}.
More recent methods have further improved these representations through a complex multi-stage learning pipeline that consists of large-scale weakly supervised contrastive training followed by multi-task fine-tuning \citep{ni-etal-2022-GTR,wang-etal-2022-e5,li-etal-2023-gte,shitao-etal-2023-bge}
Recent approaches have focused on enhancing the generalization and transferability of text embeddings using instructions \citep{su-etal-2023-Instructor,asai-etal-2023-Tart}. 

\paragraph{Unsupervised text encoders} 

Another line of work has explored training text embedders in an unsupervised manner using only a set of unordered sentences. These unsupervised approaches typically create two different representations of the same sentence for contrastive learning. The methods vary in how they form these representations -- perturbing the input sentence \citep{wu-etal-2020-clear}, or using different model instances \citep{carlsson2021semantic}. SimCSE \citep{gao-etal-2021-simcse}, the approach used in this work, generates two representations of the same sentence by passing it through the model twice with different dropout masks.

\paragraph{Turning decoder-only LLMs into text encoders} 

While decoder-only LLMs have outperformed bidirectional encoders across a large variety of language understanding tasks \citep[\textit{inter alia}]{brown-etal-gpt3, Hugo2023-llama-2,jiang2023-mistral}, their impact on sentence representation learning remains limited. 
The most common approaches in literature use the final hidden state of the last token as the sentence embedding \citep{neelakantan-2022-openai,Ma2023-repllama, Wang2023-e5mistral}.

There are few works that explore the limitations of using a causal attention mask when adapting decoder-only LLMs for text classification and sentence representation tasks. \citet{li-etal-2023-labelsupervisedllamafinetuning} experiment with removing the causal mask of Llama-2 during supervised fine-tuning for text classification and NER tasks. Similarly, \citet{dukić-etal-2024-lookingright} enable bidirectional attention for a group of layers during supervised fine-tuning on NER and chunking. In the context of sentence representation learning, \citet{li-li-2024-bellm} explore enabling bidirectional attention in the last layer of a decoder-only model during supervised contrastive fine-tuning on STS tasks. 

Concurrent to our work, several works have focused on converting decoder-only-LLMs to text encoders in supervised and unsupervised manner. \citet{jiang-etal-2023-prompteol} and \citet{lei-etal-2024-metaprompteol} prompt the language model to summarize the input text in one word, and take the last layer’s hidden embedding for the last token as the text's representation. 
\citet{muennighoff2024generative} perform multi-task full fine-tuning using a combination of self-supervised language modeling with causal attention and supervised contrastive learning with bidirectional attention. In contrast, our proposed approach is much more computationally efficient, as it requires only parameter-efficient fine-tuning and 1000 gradient steps.
Closest to our work is the concurrent work of \citet{springer2024repetition}. They propose to copy the input sequence and append it to itself, which addresses the contextualization issue of causal attention as tokens in the copy of the input can now attend to "future" tokens in the previous sequence. While this performs well in practice, it significantly increases the computational cost at inference time, which can be particularly problematic for encoding longer documents. Our approach outperforms \citet{springer2024repetition}, without inducing any additional computational overhead at inference time.

\section{Conclusion}
\label{sec:conclusion}

We present LLM2Vec, a strong unsupervised approach to transform any decoder-only LLMs into a (universal) text embedder. We perform an extensive evaluation on word- and sequence-level tasks and demonstrate the effectiveness of LLM2Vec in both unsupervised and supervised settings.
Applying LLM2Vec to \mistral{} achieves a new state-of-the-art performance on MTEB among unsupervised approaches. When combining LLM2Vec with supervised contrastive fine-tuning, \llamathree{} achieves SOTA performance among approaches that train only on publicly available data (as of May 24, 2024). 
Beyond our strong empirical contributions, we provide an extensive analysis of how LLM2Vec impacts the underlying model and reveal an intriguing property of \mistral{}, which explains its strong out of the box performance with bidirectional attention. The simplicity of our approach, as well as its compute and sample-efficiency, makes LLM2vec a promising solution for low-resource and compute constrained scenarios and opens up several interesting avenues for future work.
\section*{Acknowledgements}
We thank the members of SR’s research group for providing feedback throughout the project.
Furthermore, we thank Jacob Mitchell Springer for providing the supervised training data used in \citet{springer2024repetition}. 
PB is supported by the Mila-Intel Grant program. MM is partly funded by the Mila P2v5 Technology Maturation Grant and the Mila-Samsung grant. SR is supported by a Facebook CIFAR AI Chair and NSERC Discovery Grant program.



\bibliography{bib}

\begin{thebibliography}{59}
\providecommand{\natexlab}[1]{#1}
\providecommand{\url}[1]{\texttt{#1}}
\expandafter\ifx\csname urlstyle\endcsname\relax
  \providecommand{\doi}[1]{doi: #1}\else
  \providecommand{\doi}{doi: \begingroup \urlstyle{rm}\Url}\fi

\bibitem[Agirre et~al.(2014)Agirre, Banea, Cardie, Cer, Diab, Gonzalez-Agirre, Guo, Mihalcea, Rigau, and Wiebe]{agirre-etal-2014-SICKR}
Eneko Agirre, Carmen Banea, Claire Cardie, Daniel Cer, Mona Diab, Aitor Gonzalez-Agirre, Weiwei Guo, Rada Mihalcea, German Rigau, and Janyce Wiebe.
\newblock {S}em{E}val-2014 task 10: Multilingual semantic textual similarity.
\newblock In Preslav Nakov and Torsten Zesch (eds.), \emph{Proceedings of the 8th International Workshop on Semantic Evaluation ({S}em{E}val 2014)}, pp.\  81--91, Dublin, Ireland, August 2014. Association for Computational Linguistics.
\newblock \doi{10.3115/v1/S14-2010}.
\newblock URL \url{https://aclanthology.org/S14-2010}.

\bibitem[AI@Meta(2024)]{llama3}
AI@Meta.
\newblock Llama 3 model card.
\newblock 2024.
\newblock URL \url{https://github.com/meta-llama/llama3/blob/main/MODEL_CARD.md}.

\bibitem[Asai et~al.(2023)Asai, Schick, Lewis, Chen, Izacard, Riedel, Hajishirzi, and Yih]{asai-etal-2023-Tart}
Akari Asai, Timo Schick, Patrick Lewis, Xilun Chen, Gautier Izacard, Sebastian Riedel, Hannaneh Hajishirzi, and Wen-tau Yih.
\newblock Task-aware retrieval with instructions.
\newblock In Anna Rogers, Jordan Boyd-Graber, and Naoaki Okazaki (eds.), \emph{Findings of the Association for Computational Linguistics: ACL 2023}, pp.\  3650--3675, Toronto, Canada, July 2023. Association for Computational Linguistics.
\newblock \doi{10.18653/v1/2023.findings-acl.225}.
\newblock URL \url{https://aclanthology.org/2023.findings-acl.225}.

\bibitem[Belinkov(2022)]{belinkov-2022-probing}
Yonatan Belinkov.
\newblock Probing classifiers: Promises, shortcomings, and advances.
\newblock \emph{Computational Linguistics}, 48\penalty0 (1):\penalty0 207--219, March 2022.
\newblock \doi{10.1162/coli_a_00422}.
\newblock URL \url{https://aclanthology.org/2022.cl-1.7}.

\bibitem[Brown et~al.(2020)Brown, Mann, Ryder, Subbiah, Kaplan, Dhariwal, Neelakantan, Shyam, Sastry, Askell, Agarwal, Herbert-Voss, Krueger, Henighan, Child, Ramesh, Ziegler, Wu, Winter, Hesse, Chen, Sigler, Litwin, Gray, Chess, Clark, Berner, McCandlish, Radford, Sutskever, and Amodei]{brown-etal-gpt3}
Tom Brown, Benjamin Mann, Nick Ryder, Melanie Subbiah, Jared~D Kaplan, Prafulla Dhariwal, Arvind Neelakantan, Pranav Shyam, Girish Sastry, Amanda Askell, Sandhini Agarwal, Ariel Herbert-Voss, Gretchen Krueger, Tom Henighan, Rewon Child, Aditya Ramesh, Daniel Ziegler, Jeffrey Wu, Clemens Winter, Chris Hesse, Mark Chen, Eric Sigler, Mateusz Litwin, Scott Gray, Benjamin Chess, Jack Clark, Christopher Berner, Sam McCandlish, Alec Radford, Ilya Sutskever, and Dario Amodei.
\newblock Language models are few-shot learners.
\newblock In H.~Larochelle, M.~Ranzato, R.~Hadsell, M.F. Balcan, and H.~Lin (eds.), \emph{Advances in Neural Information Processing Systems}, volume~33, pp.\  1877--1901. Curran Associates, Inc., 2020.
\newblock URL \url{https://proceedings.neurips.cc/paper_files/paper/2020/file/1457c0d6bfcb4967418bfb8ac142f64a-Paper.pdf}.

\bibitem[Carlsson et~al.(2021)Carlsson, Gyllensten, Gogoulou, Hellqvist, and Sahlgren]{carlsson2021semantic}
Fredrik Carlsson, Amaru~Cuba Gyllensten, Evangelia Gogoulou, Erik~Ylip{\"a}{\"a} Hellqvist, and Magnus Sahlgren.
\newblock Semantic re-tuning with contrastive tension.
\newblock In \emph{International Conference on Learning Representations}, 2021.
\newblock URL \url{https://openreview.net/forum?id=Ov_sMNau-PF}.

\bibitem[Cheng et~al.(2016)Cheng, Dong, and Lapata]{cheng-etal-2016-long}
Jianpeng Cheng, Li~Dong, and Mirella Lapata.
\newblock Long short-term memory-networks for machine reading.
\newblock In Jian Su, Kevin Duh, and Xavier Carreras (eds.), \emph{Proceedings of the 2016 Conference on Empirical Methods in Natural Language Processing}, pp.\  551--561, Austin, Texas, November 2016. Association for Computational Linguistics.
\newblock \doi{10.18653/v1/D16-1053}.
\newblock URL \url{https://aclanthology.org/D16-1053}.

\bibitem[Chowdhery et~al.(2023)Chowdhery, Narang, Devlin, Bosma, Mishra, Roberts, Barham, Chung, Sutton, Gehrmann, Schuh, Shi, Tsvyashchenko, Maynez, Rao, Barnes, Tay, Shazeer, Prabhakaran, Reif, Du, Hutchinson, Pope, Bradbury, Austin, Isard, Gur-Ari, Yin, Duke, Levskaya, Ghemawat, Dev, Michalewski, Garcia, Misra, Robinson, Fedus, Zhou, Ippolito, Luan, Lim, Zoph, Spiridonov, Sepassi, Dohan, Agrawal, Omernick, Dai, Pillai, Pellat, Lewkowycz, Moreira, Child, Polozov, Lee, Zhou, Wang, Saeta, Diaz, Firat, Catasta, Wei, Meier-Hellstern, Eck, Dean, Petrov, and Fiedel]{chowdhery2022palm}
Aakanksha Chowdhery, Sharan Narang, Jacob Devlin, Maarten Bosma, Gaurav Mishra, Adam Roberts, Paul Barham, Hyung~Won Chung, Charles Sutton, Sebastian Gehrmann, Parker Schuh, Kensen Shi, Sasha Tsvyashchenko, Joshua Maynez, Abhishek Rao, Parker Barnes, Yi~Tay, Noam Shazeer, Vinodkumar Prabhakaran, Emily Reif, Nan Du, Ben Hutchinson, Reiner Pope, James Bradbury, Jacob Austin, Michael Isard, Guy Gur-Ari, Pengcheng Yin, Toju Duke, Anselm Levskaya, Sanjay Ghemawat, Sunipa Dev, Henryk Michalewski, Xavier Garcia, Vedant Misra, Kevin Robinson, Liam Fedus, Denny Zhou, Daphne Ippolito, David Luan, Hyeontaek Lim, Barret Zoph, Alexander Spiridonov, Ryan Sepassi, David Dohan, Shivani Agrawal, Mark Omernick, Andrew~M. Dai, Thanumalayan~Sankaranarayana Pillai, Marie Pellat, Aitor Lewkowycz, Erica Moreira, Rewon Child, Oleksandr Polozov, Katherine Lee, Zongwei Zhou, Xuezhi Wang, Brennan Saeta, Mark Diaz, Orhan Firat, Michele Catasta, Jason Wei, Kathy Meier-Hellstern, Douglas Eck, Jeff Dean, Slav Petrov, and Noah Fiedel.
\newblock {PaLM}: {S}caling language modeling with pathways.
\newblock \emph{Journal of Machine Learning Research}, 24\penalty0 (240):\penalty0 1--113, 2023.
\newblock URL \url{http://jmlr.org/papers/v24/22-1144.html}.

\bibitem[Clark et~al.(2020)Clark, Luong, Le, and Manning]{clark2020electra}
Kevin Clark, Minh-Thang Luong, Quoc~V. Le, and Christopher~D. Manning.
\newblock Electra: Pre-training text encoders as discriminators rather than generators.
\newblock In \emph{International Conference on Learning Representations}, 2020.
\newblock URL \url{https://openreview.net/forum?id=r1xMH1BtvB}.

\bibitem[Conneau et~al.(2017)Conneau, Kiela, Schwenk, Barrault, and Bordes]{conneau-etal-2017-supervised}
Alexis Conneau, Douwe Kiela, Holger Schwenk, Lo{\"\i}c Barrault, and Antoine Bordes.
\newblock Supervised learning of universal sentence representations from natural language inference data.
\newblock In Martha Palmer, Rebecca Hwa, and Sebastian Riedel (eds.), \emph{Proceedings of the 2017 Conference on Empirical Methods in Natural Language Processing}, pp.\  670--680, Copenhagen, Denmark, September 2017. Association for Computational Linguistics.
\newblock \doi{10.18653/v1/D17-1070}.
\newblock URL \url{https://aclanthology.org/D17-1070}.

\bibitem[Dao(2024)]{dao2023flashattention2}
Tri Dao.
\newblock {FlashAttention-2}: {F}aster attention with better parallelism and work partitioning.
\newblock In \emph{The Twelfth International Conference on Learning Representations}, 2024.
\newblock URL \url{https://openreview.net/forum?id=mZn2Xyh9Ec}.

\bibitem[DataCanary et~al.(2017)DataCanary, Lili, Meg, Nikhil, and tomtung]{quora-question-pairs}
hilfialkaff DataCanary, Jiang Lili, Risdal Meg, Dandekar Nikhil, and tomtung.
\newblock Quora question pairs.
\newblock 2017.
\newblock URL \url{https://kaggle.com/competitions/quora-question-pairs}.

\bibitem[Devlin et~al.(2019)Devlin, Chang, Lee, and Toutanova]{devlin-2019-BERT}
Jacob Devlin, Ming-Wei Chang, Kenton Lee, and Kristina Toutanova.
\newblock {BERT}: Pre-training of deep bidirectional transformers for language understanding.
\newblock In Jill Burstein, Christy Doran, and Thamar Solorio (eds.), \emph{Proceedings of the 2019 Conference of the North {A}merican Chapter of the Association for Computational Linguistics: Human Language Technologies, Volume 1 (Long and Short Papers)}, pp.\  4171--4186, Minneapolis, Minnesota, June 2019. Association for Computational Linguistics.
\newblock \doi{10.18653/v1/N19-1423}.
\newblock URL \url{https://aclanthology.org/N19-1423}.

\bibitem[Dukić \& Šnajder(2024)Dukić and Šnajder]{dukić-etal-2024-lookingright}
David Dukić and Jan Šnajder.
\newblock Looking right is sometimes right: Investigating the capabilities of decoder-only llms for sequence labeling.
\newblock \emph{arXiv preprint}, 2024.
\newblock URL \url{https://arxiv.org/abs/2401.14556}.

\bibitem[Fan et~al.(2019)Fan, Jernite, Perez, Grangier, Weston, and Auli]{fan-etal-2019-eli5}
Angela Fan, Yacine Jernite, Ethan Perez, David Grangier, Jason Weston, and Michael Auli.
\newblock {ELI}5: Long form question answering.
\newblock In Anna Korhonen, David Traum, and Llu{\'\i}s M{\`a}rquez (eds.), \emph{Proceedings of the 57th Annual Meeting of the Association for Computational Linguistics}, pp.\  3558--3567, Florence, Italy, July 2019. Association for Computational Linguistics.
\newblock \doi{10.18653/v1/P19-1346}.
\newblock URL \url{https://aclanthology.org/P19-1346}.

\bibitem[Gao et~al.(2021)Gao, Yao, and Chen]{gao-etal-2021-simcse}
Tianyu Gao, Xingcheng Yao, and Danqi Chen.
\newblock {S}im{CSE}: Simple contrastive learning of sentence embeddings.
\newblock In Marie-Francine Moens, Xuanjing Huang, Lucia Specia, and Scott Wen-tau Yih (eds.), \emph{Proceedings of the 2021 Conference on Empirical Methods in Natural Language Processing}, pp.\  6894--6910, Online and Punta Cana, Dominican Republic, November 2021. Association for Computational Linguistics.
\newblock \doi{10.18653/v1/2021.emnlp-main.552}.
\newblock URL \url{https://aclanthology.org/2021.emnlp-main.552}.

\bibitem[He et~al.(2023)He, Gao, and Chen]{he2023debertav}
Pengcheng He, Jianfeng Gao, and Weizhu Chen.
\newblock De{BERT}av3: Improving de{BERT}a using {ELECTRA}-style pre-training with gradient-disentangled embedding sharing.
\newblock In \emph{The Eleventh International Conference on Learning Representations}, 2023.
\newblock URL \url{https://openreview.net/forum?id=sE7-XhLxHA}.

\bibitem[He et~al.(2018)He, Liu, Liu, Lyu, Zhao, Xiao, Liu, Wang, Wu, She, Liu, Wu, and Wang]{he-etal-2018-dureader}
Wei He, Kai Liu, Jing Liu, Yajuan Lyu, Shiqi Zhao, Xinyan Xiao, Yuan Liu, Yizhong Wang, Hua Wu, Qiaoqiao She, Xuan Liu, Tian Wu, and Haifeng Wang.
\newblock {D}u{R}eader: a {C}hinese machine reading comprehension dataset from real-world applications.
\newblock In Eunsol Choi, Minjoon Seo, Danqi Chen, Robin Jia, and Jonathan Berant (eds.), \emph{Proceedings of the Workshop on Machine Reading for Question Answering}, pp.\  37--46, Melbourne, Australia, July 2018. Association for Computational Linguistics.
\newblock \doi{10.18653/v1/W18-2605}.
\newblock URL \url{https://aclanthology.org/W18-2605}.

\bibitem[Hu et~al.(2022)Hu, yelong shen, Wallis, Allen-Zhu, Li, Wang, Wang, and Chen]{hu2022lora}
Edward~J Hu, yelong shen, Phillip Wallis, Zeyuan Allen-Zhu, Yuanzhi Li, Shean Wang, Lu~Wang, and Weizhu Chen.
\newblock Lo{RA}: Low-rank adaptation of large language models.
\newblock In \emph{International Conference on Learning Representations}, 2022.
\newblock URL \url{https://openreview.net/forum?id=nZeVKeeFYf9}.

\bibitem[Jiang et~al.(2023{\natexlab{a}})Jiang, Sablayrolles, Mensch, Bamford, Chaplot, de~las Casas, Bressand, Lengyel, Lample, Saulnier, Lavaud, Lachaux, Stock, Scao, Lavril, Wang, Lacroix, and Sayed]{jiang2023-mistral}
Albert~Q. Jiang, Alexandre Sablayrolles, Arthur Mensch, Chris Bamford, Devendra~Singh Chaplot, Diego de~las Casas, Florian Bressand, Gianna Lengyel, Guillaume Lample, Lucile Saulnier, Lélio~Renard Lavaud, Marie-Anne Lachaux, Pierre Stock, Teven~Le Scao, Thibaut Lavril, Thomas Wang, Timothée Lacroix, and William~El Sayed.
\newblock Mistral 7{B}.
\newblock \emph{arXiv preprint}, 2023{\natexlab{a}}.
\newblock URL \url{https://arxiv.org/abs/2310.06825}.

\bibitem[Jiang et~al.(2023{\natexlab{b}})Jiang, Huang, Luan, Wang, and Zhuang]{jiang-etal-2023-prompteol}
Ting Jiang, Shaohan Huang, Zhongzhi Luan, Deqing Wang, and Fuzhen Zhuang.
\newblock Scaling sentence embeddings with large language models.
\newblock 2023{\natexlab{b}}.
\newblock URL \url{https://arxiv.org/abs/2307.16645}.

\bibitem[Joshi et~al.(2017)Joshi, Choi, Weld, and Zettlemoyer]{joshi-etal-2017-triviaqa}
Mandar Joshi, Eunsol Choi, Daniel Weld, and Luke Zettlemoyer.
\newblock {T}rivia{QA}: A large scale distantly supervised challenge dataset for reading comprehension.
\newblock In Regina Barzilay and Min-Yen Kan (eds.), \emph{Proceedings of the 55th Annual Meeting of the Association for Computational Linguistics (Volume 1: Long Papers)}, pp.\  1601--1611, Vancouver, Canada, July 2017. Association for Computational Linguistics.
\newblock \doi{10.18653/v1/P17-1147}.
\newblock URL \url{https://aclanthology.org/P17-1147}.

\bibitem[Karpukhin et~al.(2020)Karpukhin, Oguz, Min, Lewis, Wu, Edunov, Chen, and Yih]{karpukhin-etal-2020-DPR}
Vladimir Karpukhin, Barlas Oguz, Sewon Min, Patrick Lewis, Ledell Wu, Sergey Edunov, Danqi Chen, and Wen-tau Yih.
\newblock Dense passage retrieval for open-domain question answering.
\newblock In Bonnie Webber, Trevor Cohn, Yulan He, and Yang Liu (eds.), \emph{Proceedings of the 2020 Conference on Empirical Methods in Natural Language Processing (EMNLP)}, pp.\  6769--6781, Online, November 2020. Association for Computational Linguistics.
\newblock \doi{10.18653/v1/2020.emnlp-main.550}.
\newblock URL \url{https://aclanthology.org/2020.emnlp-main.550}.

\bibitem[Khattab \& Zaharia(2020)Khattab and Zaharia]{khattab-2020-ColBERT}
Omar Khattab and Matei Zaharia.
\newblock {ColBERT}: {E}fficient and effective passage search via contextualized late interaction over bert.
\newblock In \emph{Proceedings of the 43rd International ACM SIGIR Conference on Research and Development in Information Retrieval}, SIGIR '20, pp.\  39–48, New York, NY, USA, 2020. Association for Computing Machinery.
\newblock ISBN 9781450380164.
\newblock \doi{10.1145/3397271.3401075}.
\newblock URL \url{https://doi.org/10.1145/3397271.3401075}.

\bibitem[Lei et~al.(2024)Lei, Wu, Zhou, Shen, Cao, Tao, and Yates]{lei-etal-2024-metaprompteol}
Yibin Lei, Di~Wu, Tianyi Zhou, Tao Shen, Yu~Cao, Chongyang Tao, and Andrew Yates.
\newblock Meta-task prompting elicits embeddings from large language models.
\newblock 2024.
\newblock URL \url{https://arxiv.org/abs/2402.18458}.

\bibitem[Li \& Li(2024)Li and Li]{li-li-2024-bellm}
Xianming Li and Jing Li.
\newblock {B}e{LLM}: Backward dependency enhanced large language model for sentence embeddings.
\newblock In Kevin Duh, Helena Gomez, and Steven Bethard (eds.), \emph{Proceedings of the 2024 Conference of the North American Chapter of the Association for Computational Linguistics: Human Language Technologies (Volume 1: Long Papers)}, pp.\  792--804, Mexico City, Mexico, June 2024. Association for Computational Linguistics.
\newblock \doi{10.18653/v1/2024.naacl-long.45}.
\newblock URL \url{https://aclanthology.org/2024.naacl-long.45}.

\bibitem[Li et~al.(2023{\natexlab{a}})Li, Zhang, Zhang, Long, Xie, and Zhang]{li-etal-2023-gte}
Zehan Li, Xin Zhang, Yanzhao Zhang, Dingkun Long, Pengjun Xie, and Meishan Zhang.
\newblock Towards general text embeddings with multi-stage contrastive learning.
\newblock \emph{arXiv preprint}, 2023{\natexlab{a}}.
\newblock URL \url{https://arxiv.org/abs/2308.03281}.

\bibitem[Li et~al.(2023{\natexlab{b}})Li, Li, Liu, Xie, Li, lee Wang, Li, and Zhong]{li-etal-2023-labelsupervisedllamafinetuning}
Zongxi Li, Xianming Li, Yuzhang Liu, Haoran Xie, Jing Li, Fu~lee Wang, Qing Li, and Xiaoqin Zhong.
\newblock Label supervised llama finetuning.
\newblock \emph{arXiv preprint}, 2023{\natexlab{b}}.
\newblock URL \url{https://arxiv.org/abs/2310.01208}.

\bibitem[Lin et~al.(2017)Lin, Feng, dos Santos, Yu, Xiang, Zhou, and Bengio]{lin2017a}
Zhouhan Lin, Minwei Feng, Cicero~Nogueira dos Santos, Mo~Yu, Bing Xiang, Bowen Zhou, and Yoshua Bengio.
\newblock A structured self-attentive sentence embedding.
\newblock In \emph{International Conference on Learning Representations}, 2017.
\newblock URL \url{https://openreview.net/forum?id=BJC_jUqxe}.

\bibitem[Liu et~al.(2019)Liu, Ott, Goyal, Du, Joshi, Chen, Levy, Lewis, Zettlemoyer, and Stoyanov]{liu2019-roberta}
Yinhan Liu, Myle Ott, Naman Goyal, Jingfei Du, Mandar Joshi, Danqi Chen, Omer Levy, Mike Lewis, Luke Zettlemoyer, and Veselin Stoyanov.
\newblock {RoBERTa:} {A} robustly optimized {BERT} pretraining approach.
\newblock \emph{arXiv preprint}, 2019.
\newblock URL \url{http://arxiv.org/abs/1907.11692}.

\bibitem[Lv et~al.(2023)Lv, Zhang, Xie, Tu, Chen, Wen, and Yan]{lv-etal-2023-falling}
Ang Lv, Kaiyi Zhang, Shufang Xie, Quan Tu, Yuhan Chen, Ji-Rong Wen, and Rui Yan.
\newblock Are we falling in a middle-intelligence trap? an analysis and mitigation of the reversal curse.
\newblock \emph{arXiv preprint}, 2023.
\newblock URL \url{https://arxiv.org/abs/2311.07468}.

\bibitem[Ma et~al.(2023)Ma, Wang, Yang, Wei, and Lin]{Ma2023-repllama}
Xueguang Ma, Liang Wang, Nan Yang, Furu Wei, and Jimmy Lin.
\newblock Fine-tuning {LLaMA} for multi-stage text retrieval.
\newblock \emph{arXiv preprint}, 2023.
\newblock URL \url{https://arxiv.org/abs/2310.08319}.

\bibitem[Merity et~al.(2017)Merity, Xiong, Bradbury, and Socher]{merity2017pointer}
Stephen Merity, Caiming Xiong, James Bradbury, and Richard Socher.
\newblock Pointer sentinel mixture models.
\newblock In \emph{International Conference on Learning Representations}, 2017.
\newblock URL \url{https://openreview.net/forum?id=Byj72udxe}.

\bibitem[Muennighoff(2022)]{Muennighoff2022-SGPT}
Niklas Muennighoff.
\newblock {SGPT}: {GPT} sentence embeddings for semantic search.
\newblock \emph{arXiv preprint}, 2022.
\newblock URL \url{https://arxiv.org/abs/2202.08904}.

\bibitem[Muennighoff et~al.(2023)Muennighoff, Tazi, Magne, and Reimers]{muennighoff-etal-2023-mteb}
Niklas Muennighoff, Nouamane Tazi, Loic Magne, and Nils Reimers.
\newblock {MTEB}: Massive text embedding benchmark.
\newblock In Andreas Vlachos and Isabelle Augenstein (eds.), \emph{Proceedings of the 17th Conference of the European Chapter of the Association for Computational Linguistics}, pp.\  2014--2037, Dubrovnik, Croatia, May 2023. Association for Computational Linguistics.
\newblock \doi{10.18653/v1/2023.eacl-main.148}.
\newblock URL \url{https://aclanthology.org/2023.eacl-main.148}.

\bibitem[Muennighoff et~al.(2024)Muennighoff, Su, Wang, Yang, Wei, Yu, Singh, and Kiela]{muennighoff2024generative}
Niklas Muennighoff, Hongjin Su, Liang Wang, Nan Yang, Furu Wei, Tao Yu, Amanpreet Singh, and Douwe Kiela.
\newblock Generative representational instruction tuning.
\newblock \emph{arXiv preprint}, 2024.
\newblock URL \url{https://arxiv.org/abs/2402.09906}.

\bibitem[Neelakantan et~al.(2022)Neelakantan, Xu, Puri, Radford, Han, Tworek, Yuan, Tezak, Kim, Hallacy, Heidecke, Shyam, Power, Nekoul, Sastry, Krueger, Schnurr, Such, Hsu, Thompson, Khan, Sherbakov, Jang, Welinder, and Weng]{neelakantan-2022-openai}
Arvind Neelakantan, Tao Xu, Raul Puri, Alec Radford, Jesse~Michael Han, Jerry Tworek, Qiming Yuan, Nikolas Tezak, Jong~Wook Kim, Chris Hallacy, Johannes Heidecke, Pranav Shyam, Boris Power, Tyna~Eloundou Nekoul, Girish Sastry, Gretchen Krueger, David Schnurr, Felipe~Petroski Such, Kenny Hsu, Madeleine Thompson, Tabarak Khan, Toki Sherbakov, Joanne Jang, Peter Welinder, and Lilian Weng.
\newblock Text and code embeddings by contrastive pre-training.
\newblock \emph{arXiv preprint}, 2022.
\newblock URL \url{https://arxiv.org/abs/2201.10005}.

\bibitem[Ni et~al.(2022)Ni, Qu, Lu, Dai, Hernandez~Abrego, Ma, Zhao, Luan, Hall, Chang, and Yang]{ni-etal-2022-GTR}
Jianmo Ni, Chen Qu, Jing Lu, Zhuyun Dai, Gustavo Hernandez~Abrego, Ji~Ma, Vincent Zhao, Yi~Luan, Keith Hall, Ming-Wei Chang, and Yinfei Yang.
\newblock Large dual encoders are generalizable retrievers.
\newblock In Yoav Goldberg, Zornitsa Kozareva, and Yue Zhang (eds.), \emph{Proceedings of the 2022 Conference on Empirical Methods in Natural Language Processing}, pp.\  9844--9855, Abu Dhabi, United Arab Emirates, December 2022. Association for Computational Linguistics.
\newblock \doi{10.18653/v1/2022.emnlp-main.669}.
\newblock URL \url{https://aclanthology.org/2022.emnlp-main.669}.

\bibitem[Ouyang et~al.(2022)Ouyang, Wu, Jiang, Almeida, Wainwright, Mishkin, Zhang, Agarwal, Slama, Ray, Schulman, Hilton, Kelton, Miller, Simens, Askell, Welinder, Christiano, Leike, and Lowe]{ouyang2022training}
Long Ouyang, Jeff Wu, Xu~Jiang, Diogo Almeida, Carroll~L. Wainwright, Pamela Mishkin, Chong Zhang, Sandhini Agarwal, Katarina Slama, Alex Ray, John Schulman, Jacob Hilton, Fraser Kelton, Luke Miller, Maddie Simens, Amanda Askell, Peter Welinder, Paul Christiano, Jan Leike, and Ryan Lowe.
\newblock Training language models to follow instructions with human feedback.
\newblock \emph{arXiv preprint}, 2022.
\newblock URL \url{https://arxiv.org/abs/2203.02155}.

\bibitem[Paulus et~al.(2018)Paulus, Xiong, and Socher]{paulus2018a}
Romain Paulus, Caiming Xiong, and Richard Socher.
\newblock A deep reinforced model for abstractive summarization.
\newblock In \emph{International Conference on Learning Representations}, 2018.
\newblock URL \url{https://openreview.net/forum?id=HkAClQgA-}.

\bibitem[Raffel et~al.(2020)Raffel, Shazeer, Roberts, Lee, Narang, Matena, Zhou, Li, and Liu]{raffel-etal-2020-t5}
Colin Raffel, Noam Shazeer, Adam Roberts, Katherine Lee, Sharan Narang, Michael Matena, Yanqi Zhou, Wei Li, and Peter~J. Liu.
\newblock Exploring the limits of transfer learning with a unified text-to-text transformer.
\newblock \emph{Journal of Machine Learning Research}, 21\penalty0 (140):\penalty0 1--67, 2020.
\newblock URL \url{http://jmlr.org/papers/v21/20-074.html}.

\bibitem[Rajpurkar et~al.(2016)Rajpurkar, Zhang, Lopyrev, and Liang]{rajpurkar-etal-2016-squad}
Pranav Rajpurkar, Jian Zhang, Konstantin Lopyrev, and Percy Liang.
\newblock {SQ}u{AD}: 100,000+ questions for machine comprehension of text.
\newblock In Jian Su, Kevin Duh, and Xavier Carreras (eds.), \emph{Proceedings of the 2016 Conference on Empirical Methods in Natural Language Processing}, pp.\  2383--2392, Austin, Texas, November 2016. Association for Computational Linguistics.
\newblock \doi{10.18653/v1/D16-1264}.
\newblock URL \url{https://aclanthology.org/D16-1264}.

\bibitem[Reimers \& Gurevych(2019)Reimers and Gurevych]{reimers-gurevych-2019-sbert}
Nils Reimers and Iryna Gurevych.
\newblock Sentence-{BERT}: Sentence embeddings using {S}iamese {BERT}-networks.
\newblock In Kentaro Inui, Jing Jiang, Vincent Ng, and Xiaojun Wan (eds.), \emph{Proceedings of the 2019 Conference on Empirical Methods in Natural Language Processing and the 9th International Joint Conference on Natural Language Processing (EMNLP-IJCNLP)}, pp.\  3982--3992, Hong Kong, China, November 2019. Association for Computational Linguistics.
\newblock \doi{10.18653/v1/D19-1410}.
\newblock URL \url{https://aclanthology.org/D19-1410}.

\bibitem[Springer et~al.(2024)Springer, Kotha, Fried, Neubig, and Raghunathan]{springer2024repetition}
Jacob~Mitchell Springer, Suhas Kotha, Daniel Fried, Graham Neubig, and Aditi Raghunathan.
\newblock Repetition improves language model embeddings.
\newblock \emph{arXiv preprint}, 2024.
\newblock URL \url{https://arxiv.org/abs/2402.15449}.

\bibitem[Su et~al.(2023)Su, Shi, Kasai, Wang, Hu, Ostendorf, Yih, Smith, Zettlemoyer, and Yu]{su-etal-2023-Instructor}
Hongjin Su, Weijia Shi, Jungo Kasai, Yizhong Wang, Yushi Hu, Mari Ostendorf, Wen-tau Yih, Noah~A. Smith, Luke Zettlemoyer, and Tao Yu.
\newblock One embedder, any task: Instruction-finetuned text embeddings.
\newblock In Anna Rogers, Jordan Boyd-Graber, and Naoaki Okazaki (eds.), \emph{Findings of the Association for Computational Linguistics: ACL 2023}, pp.\  1102--1121, Toronto, Canada, July 2023. Association for Computational Linguistics.
\newblock \doi{10.18653/v1/2023.findings-acl.71}.
\newblock URL \url{https://aclanthology.org/2023.findings-acl.71}.

\bibitem[Thorne et~al.(2018)Thorne, Vlachos, Christodoulopoulos, and Mittal]{thorne-etal-2018-fever}
James Thorne, Andreas Vlachos, Christos Christodoulopoulos, and Arpit Mittal.
\newblock {FEVER}: {A} large-scale dataset for fact extraction and {VER}ification.
\newblock In Marilyn Walker, Heng Ji, and Amanda Stent (eds.), \emph{Proceedings of the 2018 Conference of the North {A}merican Chapter of the Association for Computational Linguistics: Human Language Technologies, Volume 1 (Long Papers)}, pp.\  809--819, New Orleans, Louisiana, June 2018. Association for Computational Linguistics.
\newblock \doi{10.18653/v1/N18-1074}.
\newblock URL \url{https://aclanthology.org/N18-1074}.

\bibitem[Tjong Kim~Sang \& De~Meulder(2003)Tjong Kim~Sang and De~Meulder]{conll-2003}
Erik~F. Tjong Kim~Sang and Fien De~Meulder.
\newblock Introduction to the {C}o{NLL}-2003 shared task: Language-independent named entity recognition.
\newblock In \emph{Proceedings of the Seventh Conference on Natural Language Learning at {HLT}-{NAACL} 2003}, pp.\  142--147, 2003.
\newblock URL \url{https://www.aclweb.org/anthology/W03-0419}.

\bibitem[Touvron et~al.(2023)Touvron, Martin, Stone, Albert, Almahairi, Babaei, Bashlykov, Batra, Bhargava, Bhosale, Bikel, Blecher, Ferrer, Chen, Cucurull, Esiobu, Fernandes, Fu, Fu, Fuller, Gao, Goswami, Goyal, Hartshorn, Hosseini, Hou, Inan, Kardas, Kerkez, Khabsa, Kloumann, Korenev, Koura, Lachaux, Lavril, Lee, Liskovich, Lu, Mao, Martinet, Mihaylov, Mishra, Molybog, Nie, Poulton, Reizenstein, Rungta, Saladi, Schelten, Silva, Smith, Subramanian, Tan, Tang, Taylor, Williams, Kuan, Xu, Yan, Zarov, Zhang, Fan, Kambadur, Narang, Rodriguez, Stojnic, Edunov, and Scialom]{Hugo2023-llama-2}
Hugo Touvron, Louis Martin, Kevin~R. Stone, Peter Albert, Amjad Almahairi, Yasmine Babaei, Nikolay Bashlykov, Soumya Batra, Prajjwal Bhargava, Shruti Bhosale, D.~Bikel, Lukas Blecher, Cristian~Cantón Ferrer, Moya Chen, Guillem Cucurull, David Esiobu, Jude Fernandes, Jeremy Fu, Wenyin Fu, Brian Fuller, Cynthia Gao, Vedanuj Goswami, Naman Goyal, A.~Hartshorn, Saghar Hosseini, Rui Hou, Hakan Inan, Marcin Kardas, Viktor Kerkez, Madian Khabsa, Isabel~M. Kloumann, A.~Korenev, Punit~Singh Koura, Marie-Anne Lachaux, Thibaut Lavril, Jenya Lee, Diana Liskovich, Yinghai Lu, Yuning Mao, Xavier Martinet, Todor Mihaylov, Pushkar Mishra, Igor Molybog, Yixin Nie, Andrew Poulton, Jeremy Reizenstein, Rashi Rungta, Kalyan Saladi, Alan Schelten, Ruan Silva, Eric~Michael Smith, R.~Subramanian, Xia Tan, Binh Tang, Ross Taylor, Adina Williams, Jian~Xiang Kuan, Puxin Xu, Zhengxu Yan, Iliyan Zarov, Yuchen Zhang, Angela Fan, Melanie Kambadur, Sharan Narang, Aurelien Rodriguez, Robert Stojnic, Sergey Edunov, and Thomas Scialom.
\newblock Llama 2: Open foundation and fine-tuned chat models.
\newblock \emph{preprint}, 2023.
\newblock URL \url{https://arxiv.org/abs/2307.09288}.

\bibitem[Vaswani et~al.(2017)Vaswani, Shazeer, Parmar, Uszkoreit, Jones, Gomez, Kaiser, and Polosukhin]{vaswani-etal-2017-Attention}
Ashish Vaswani, Noam Shazeer, Niki Parmar, Jakob Uszkoreit, Llion Jones, Aidan~N Gomez, \L~ukasz Kaiser, and Illia Polosukhin.
\newblock Attention is all you need.
\newblock In I.~Guyon, U.~Von Luxburg, S.~Bengio, H.~Wallach, R.~Fergus, S.~Vishwanathan, and R.~Garnett (eds.), \emph{Advances in Neural Information Processing Systems}, volume~30. Curran Associates, Inc., 2017.
\newblock URL \url{https://proceedings.neurips.cc/paper_files/paper/2017/file/3f5ee243547dee91fbd053c1c4a845aa-Paper.pdf}.

\bibitem[Wang et~al.(2022{\natexlab{a}})Wang, Yang, Huang, Jiao, Yang, Jiang, Majumder, and Wei]{wang-etal-2022-e5}
Liang Wang, Nan Yang, Xiaolong Huang, Binxing Jiao, Linjun Yang, Daxin Jiang, Rangan Majumder, and Furu Wei.
\newblock Text embeddings by weakly-supervised contrastive pre-training.
\newblock \emph{arXiv preprint}, 2022{\natexlab{a}}.
\newblock URL \url{https://arxiv.org/abs/2212.03533}.

\bibitem[Wang et~al.(2023)Wang, Yang, Huang, Yang, Majumder, and Wei]{Wang2023-e5mistral}
Liang Wang, Nan Yang, Xiaolong Huang, Linjun Yang, Rangan Majumder, and Furu Wei.
\newblock Improving text embeddings with large language models.
\newblock \emph{arXiv preprint}, 2023.
\newblock URL \url{https://arxiv.org/abs/2401.00368}.

\bibitem[Wang et~al.(2022{\natexlab{b}})Wang, Mishra, Alipoormolabashi, Kordi, Mirzaei, Naik, Ashok, Dhanasekaran, Arunkumar, Stap, Pathak, Karamanolakis, Lai, Purohit, Mondal, Anderson, Kuznia, Doshi, Pal, Patel, Moradshahi, Parmar, Purohit, Varshney, Kaza, Verma, Puri, Karia, Doshi, Sampat, Mishra, Reddy~A, Patro, Dixit, and Shen]{wang-etal-2022-superni}
Yizhong Wang, Swaroop Mishra, Pegah Alipoormolabashi, Yeganeh Kordi, Amirreza Mirzaei, Atharva Naik, Arjun Ashok, Arut~Selvan Dhanasekaran, Anjana Arunkumar, David Stap, Eshaan Pathak, Giannis Karamanolakis, Haizhi Lai, Ishan Purohit, Ishani Mondal, Jacob Anderson, Kirby Kuznia, Krima Doshi, Kuntal~Kumar Pal, Maitreya Patel, Mehrad Moradshahi, Mihir Parmar, Mirali Purohit, Neeraj Varshney, Phani~Rohitha Kaza, Pulkit Verma, Ravsehaj~Singh Puri, Rushang Karia, Savan Doshi, Shailaja~Keyur Sampat, Siddhartha Mishra, Sujan Reddy~A, Sumanta Patro, Tanay Dixit, and Xudong Shen.
\newblock Super-{N}atural{I}nstructions: Generalization via declarative instructions on 1600+ {NLP} tasks.
\newblock In Yoav Goldberg, Zornitsa Kozareva, and Yue Zhang (eds.), \emph{Proceedings of the 2022 Conference on Empirical Methods in Natural Language Processing}, pp.\  5085--5109, Abu Dhabi, United Arab Emirates, December 2022{\natexlab{b}}. Association for Computational Linguistics.
\newblock \doi{10.18653/v1/2022.emnlp-main.340}.
\newblock URL \url{https://aclanthology.org/2022.emnlp-main.340}.

\bibitem[Wu et~al.(2020)Wu, Wang, Gu, Khabsa, Sun, and Ma]{wu-etal-2020-clear}
Zhuofeng Wu, Sinong Wang, Jiatao Gu, Madian Khabsa, Fei Sun, and Hao Ma.
\newblock {CLEAR}: {C}ontrastive learning for sentence representation.
\newblock \emph{arXiv preprint}, 2020.
\newblock URL \url{https://arxiv.org/abs/2012.15466}.

\bibitem[Xia et~al.(2023)Xia, Gao, Zeng, and Chen]{xia2023-sheared-llama}
Mengzhou Xia, Tianyu Gao, Zhiyuan Zeng, and Danqi Chen.
\newblock Sheared {LL}a{MA}: Accelerating language model pre-training via structured pruning.
\newblock In \emph{Workshop on Advancing Neural Network Training: Computational Efficiency, Scalability, and Resource Optimization (WANT@NeurIPS 2023)}, 2023.
\newblock URL \url{https://openreview.net/forum?id=6s77hjBNfS}.

\bibitem[Xiao et~al.(2023)Xiao, Liu, Zhang, and Muennighoff]{shitao-etal-2023-bge}
Shitao Xiao, Zheng Liu, Peitian Zhang, and Niklas Muennighoff.
\newblock {C-Pack}: {P}ackaged resources to advance general chinese embedding.
\newblock \emph{arXiv preprint}, 2023.
\newblock URL \url{https://arxiv.org/abs/2309.07597}.

\bibitem[Xie et~al.(2023)Xie, Dong, Wang, Lv, Yao, Gan, Wu, Li, Li, Liu, and Ma]{t2ranking}
Xiaohui Xie, Qian Dong, Bingning Wang, Feiyang Lv, Ting Yao, Weinan Gan, Zhijing Wu, Xiangsheng Li, Haitao Li, Yiqun Liu, and Jin Ma.
\newblock T2ranking: A large-scale chinese benchmark for passage ranking.
\newblock In \emph{Proceedings of the 46th International ACM SIGIR Conference on Research and Development in Information Retrieval}, SIGIR '23, pp.\  2681–2690, New York, NY, USA, 2023. Association for Computing Machinery.
\newblock ISBN 9781450394086.
\newblock \doi{10.1145/3539618.3591874}.
\newblock URL \url{https://doi.org/10.1145/3539618.3591874}.

\bibitem[Yang et~al.(2018)Yang, Qi, Zhang, Bengio, Cohen, Salakhutdinov, and Manning]{yang-etal-2018-hotpotqa}
Zhilin Yang, Peng Qi, Saizheng Zhang, Yoshua Bengio, William Cohen, Ruslan Salakhutdinov, and Christopher~D. Manning.
\newblock {H}otpot{QA}: A dataset for diverse, explainable multi-hop question answering.
\newblock In Ellen Riloff, David Chiang, Julia Hockenmaier, and Jun{'}ichi Tsujii (eds.), \emph{Proceedings of the 2018 Conference on Empirical Methods in Natural Language Processing}, pp.\  2369--2380, Brussels, Belgium, October-November 2018. Association for Computational Linguistics.
\newblock \doi{10.18653/v1/D18-1259}.
\newblock URL \url{https://aclanthology.org/D18-1259}.

\bibitem[Zhang et~al.(2021)Zhang, Ma, Shi, and Lin]{zhang-etal-2021-mr}
Xinyu Zhang, Xueguang Ma, Peng Shi, and Jimmy Lin.
\newblock Mr. {T}y{D}i: A multi-lingual benchmark for dense retrieval.
\newblock In Duygu Ataman, Alexandra Birch, Alexis Conneau, Orhan Firat, Sebastian Ruder, and Gozde~Gul Sahin (eds.), \emph{Proceedings of the 1st Workshop on Multilingual Representation Learning}, pp.\  127--137, Punta Cana, Dominican Republic, November 2021. Association for Computational Linguistics.
\newblock \doi{10.18653/v1/2021.mrl-1.12}.
\newblock URL \url{https://aclanthology.org/2021.mrl-1.12}.

\bibitem[Zhang et~al.(2023)Zhang, Thakur, Ogundepo, Kamalloo, Alfonso-Hermelo, Li, Liu, Rezagholizadeh, and Lin]{zhang-et-all-2023-MIRACL}
Xinyu Zhang, Nandan Thakur, Odunayo Ogundepo, Ehsan Kamalloo, David Alfonso-Hermelo, Xiaoguang Li, Qun Liu, Mehdi Rezagholizadeh, and Jimmy Lin.
\newblock {MIRACL: A Multilingual Retrieval Dataset Covering 18 Diverse Languages}.
\newblock \emph{Transactions of the Association for Computational Linguistics}, 11:\penalty0 1114--1131, 09 2023.
\newblock ISSN 2307-387X.
\newblock \doi{10.1162/tacl_a_00595}.
\newblock URL \url{https://doi.org/10.1162/tacl\_a\_00595}.

\end{thebibliography}
\bibliographystyle{colm2024_conference}

\clearpage
\newpage
\appendix
\section{Limitations}
\label{sec:appendix:limitations}


\paragraph{Large size of decoder-only LLMs}

Recent years have seen a increasing trend towards training very large decoder-only LLMs, with model sizes up to 540B parameters \citep{brown-etal-gpt3,chowdhery2022palm}.
The parameter size of the model has a direct impact on the training and inference latency. Additionally, the large output embedding dimension of these models (e.g., 4096 for \mistral{} compared to 768 for \texttt{BERT}) also makes them more memory and compute intensive for creating vector indexes for large document collections.
While some of these limitations can be offset by recent advances in improving the training and inference efficiency of large models \citep{hu2022lora, dao2023flashattention2}, these techniques can technically be applied to smaller bidirectional models as well.

The advantages of small bidirectional encoders come at the cost of complex and computationally intensive training regimes \citep{li-etal-2023-gte, shitao-etal-2023-bge,li-etal-2023-gte}. In contrast, decoder-only models are much more sample-efficient and do not require large-scale contrastive pre-training \citep{Wang2023-e5mistral}. Moreover, the instruction following capabilities of decoder-only models make them strong contenders for building text embedding models that generalize to a wide range of tasks and domains without the need for expensive adaptation.

While smaller models can be more practical for some applications, the sample-efficiency, the instruction following capabilities, and the widespread use of these models in the community motivates the need to explore the potential of decoder-only LLMs for text embedding tasks.

\paragraph{Data contamination from pre-training} 

As our supervised data contains train splits of publicly available datasets, there is an extremely low chance of test set contamination with the MTEB benchmark. However there is a possibility of contamination from the pre-training data of \llama{} and \mistral{} models (\sllama{} was distilled from \llama{}). As the complete details of the pre-training data are not publicly available, we cannot be certain about the extent of contamination. However, to reliably compare with other works, we stick to our choice of model and evaluation benchmark.
We leave it to future work to investigate the performance of these models on newly designed benchmarks that are not part of their pre-training data.

\paragraph{Extending to other languages}

In this work, we have implemented and evaluated our proposed methodology -- LLM2Vec -- using only English text corpora and benchmarks. However, the methodology is language-agnostic and can be easily extended to other languages using just unstructured text collections. We leave it to future work to investigate the performance of LLM2Vec on other languages. 
\section{Background}
\label{sec:appendix:background}

\subsection{Self-attention}
\label{sec:appendix:self-attention}

The self-attention mechanism is a crucial component of decoder-only LLMs~\cite{ cheng-etal-2016-long, lin2017a, vaswani-etal-2017-Attention, paulus2018a}. 
Given a sequence of $N$ tokens, the token representation at any given transformer layer $(\mathbf{x}_1, \mathbf{x}_2, \ldots, \mathbf{x}_N)$ with $\mathbf{x}_i \in \mathbf{R}^d$ are stacked into a matrix $\mathbf{X} \in \R^{N\times d}$. Given this matrix, the self-attention mechanism computes the query, key, and value matrices $\mathbf{Q}, \mathbf{K}, \mathbf{V} \in \R^{N\times p}$ via a learned linear transformation.

\vspace{-1.5em}
\begin{align}
    \mathbf{Q} &= X \mathbf{W}^Q, \\
    \mathbf{K} &= X \mathbf{W}^K, \\
    \mathbf{V} &= X \mathbf{W}^V~.
\end{align}

The output of the self-attention layer is then computed as a linear combination of the values, weighted by the normalized inner product between keys and queries: 


\begin{align}
    \mathbf{O} = \text{softmax}\left(\frac{\mathcal{M}_{\{j\leq i\}}\mathbf{Q}\mathbf{K}^T}{\sqrt{d}}\right)\mathbf{V}~.
\end{align}

This output is then passed through a feed-forward network and added to the residual stream to obtain the token representations at the next layer. Crucially, in the case of decoder-only LLMs, the attention mask $\mathcal{M}_{\{j\leq i\}}$ prevents accessing token embeddings to the right of the current token. 


\subsection{Contrastive learning}
\label{sec:appendix:contrastive-learning}


Contrastive learning is a popular paradigm to learn text representations \cite{karpukhin-etal-2020-DPR, gao-etal-2021-simcse, su-etal-2023-Instructor, Wang2023-e5mistral, springer2024repetition}. In the supervised setup, we have a set of positive pairs $\mathcal{D}=\{(q_i, d^+_i)\}_{i=1}^n$, and a set of negative documents that can include hard or in-batch negatives. The model is trained to maximize the similarity (i.e., usually cosine similarity) of positive pairs and minimize the similarity of negative pairs, i.e., we optimize the following objective:

\begin{align}
    \mathcal{L} =  \frac{e^{\lambda s(q,d^+)}}{e^{\lambda s(q,d^+)} + \sum_{d^- \in N}{e^{\lambda s(q,d^-)}}}~,
\end{align}

where $s$ is a similarity metric, $\lambda$ a temperature value, and $N$ all the negative documents for query $q$. 

\paragraph{Unsupervised contrastive learning}
In unsupervised contrastive learning, no positive or hard negative pairs are available. Most unsupervised approaches construct two different representation for the same sample, using either model or input perturbations. SimCSE \citep{gao-etal-2021-simcse}, the unsupervised approach used in this work, creates two different representations of the same input by using independently sampled dropout masks in the intermediate model representations and train the model with in-batch negatives.






\section{Massive Text Embeddings Benchmark (MTEB)}
\label{sec:appendix:mteb-background}




\subsection{MTEB subset details}
\label{subsec:appendix:mteb}

MTEB consists of diverse small and large embedding tasks. To speed up the evaluation\footnote{Full evaluation on MTEB takes more than 40h for \mistral{} on 8x A100 GPUs.}, we consider a representative subset of 15 tasks from MTEB for our analyses, presented in \Cref{tab:appendix:mteb-subset}. 
To make sure that our ablation and analyses are not  biased towards one specific category or task, this subset includes tasks from each category with almost the same proportion compared to the full MTEB\footnote{Since the MTEB's SummEval category includes only one dataset, we skip this category in our small-scale evaluation.}. 

\begin{table}[ht]
    \centering
    \small
    \begin{tabular}{ll}
    \toprule
    \textbf{Category} & \textbf{Dataset} \\
    \midrule
    \multirow{3}{*}{Retrieval (3)} & SciFact \\
    & ArguAna \\
    & NFCorpus \\ \midrule
    \multirow{2}{*}{Reranking (2)} & StackOverflowDupQuestions \\
    & SciDocsRR\\ \midrule
    \multirow{3}{*}{Clustering (3)} & BiorxivClusteringS2S \\
    & MedrxivClusteringS2S \\
    & TwentyNewsgroupsClustering \\ \midrule
    Pair Classification (1) & SprintDuplicateQuestions \\ \midrule
    \multirow{3}{*}{Classification (3)} & Banking77Classification \\
    & EmotionClassification \\
    & MassiveIntentClassification \\ \midrule
    \multirow{3}{*}{STS (3)} & STS17 \\
    & SICK-R \\
    & STSBenchmark \\ \midrule
    SummEval (0) & -\\ \midrule
    Overall & 15 datasets \\
    \bottomrule
    \end{tabular}
    \caption{Subset of MTEB tasks used for our ablations and analysis.}\label{tab:appendix:mteb-subset}
\end{table}

\subsection{MTEB instructions}
\label{subsec:appendix:mteb-instruction}

When evaluating on MTEB, we use the same instructions as \citet{Wang2023-e5mistral}. The list of instructions for each task is listed in \Cref{tab:appendix:mteb_instructions}. 
\section{Details on unsupervised results}
\label{sec:appendix:unsupervised}

\subsection{Training details}

\subsubsection{MNTP training details}
\label{subsec:mntp-training-details}
The second step of LLM2Vec includes MNTP training. We follow established practice from the encoder-only literature for choosing our masking strategy. For example, \citep{devlin-2019-BERT} mask $15\%$ of the tokens in the input. $10\%$ of the masked tokens are then replaced with a random token from the vocabulary, while another $10\%$ are unmasked again, but still considered when computing the loss. As another example, RoBERTa \citep{liu2019-roberta} also masks $15\%$ of the input tokens but applies no further post-processing to the masked tokens. 

For our models, we perform a hyperparameter search to select the percentage of the masked tokens in a sequence choosing from $20\%, 40\%, 60\%, 80\%$, and $90\%$. For each model, we take the best setup (i.e., masking probability and BERT vs. RoBERTa approach) based on the performance on SICK-R \citep{agirre-etal-2014-SICKR} task from the MTEB dataset. This results in the following choices: for \sllama{}, \llama{}, and \llamathree{}, we apply BERT's masking strategy with masking probability of $20\%$. For \mistral{}, we apply RoBERTa's masking strategy with probability of $80\%$.\looseness-1

We train all the models for 1000 steps with LoRA $r = 16$ and $\alpha = 32$, and we follow the same training parameters as RoBERTa MNTP training. When training large 7B and 8B models, we apply brain floating point (bfloat16) quantization, as well as flash attention 2 and gradient checkpointing.

\subsubsection{SimCSE training details}
\label{subsec:simcse-training-details}

The last step of LLM2Vec involves unsupervised contrastive learning with SimCSE. Our initial experiments indicated that the low value of dropout probability ($0.1$) typically used by bidirectional encoders \citep{gao-etal-2021-simcse} does not lead to optimal performance for larger decoder-only LLMs. Therefore, we use a higher dropout probability of $0.3$ for all models.

Similar to MNTP, we train all models with LoRA $r = 16$ and $\alpha = 32$ for 1000 steps. For \llama{}, \mistral{}, and \llamathree{}, we train with a batch size of 128. For \sllama{}, we use a batch size of 32. Additionally, when training \llama{} \mistral{}, and \llamathree{}, we apply brain floating point (bfloat16) quantization, flash attention 2, and gradient checkpointing.


\subsubsection{Word-level training details}
\label{subsec:word-level-training-details}

We evaluate on three popular word embedding tasks: chunking, named-entity recognition (NER), and part-of-speech (POS) tagging. We train a linear classifier using dropout with a dropout probability of 0.1 on top of the frozen representations obtained from the last hidden layer of a model.

We use data from CoNLL-2003, consisting of roughly 14,000 training, 3,250 validation, and 3,450 test samples \citep{conll-2003}. We train the classifier for 1,500 steps with a learning rate of $5e-4$ and a batch size of 8. For experiments with \mistral{} models that have been tuned with MNTP, we use the variant which is trained with BERT's masking strategy and masking probability of 20\% (please see \ref{subsec:mntp-training-details} for more details). Although 80\% masking helps with the performance in sentence-level tasks, it prevents the model from learning proper token representations essential for word-level tasks. 

Since the models we experiment with have sub-token based vocabularies, we calculate the embedding of a word by averaging the representations of all its sub-tokens. For example, for a sentence ``$w_1\; w_2\; w_3$'' which is tokenized as ``$\text{BOS} \;\; t_{11} t_{12} \;\; t_{21} t_{22} t_{23} \;\; t_{31}$'', the representation of $w_1$, $w_2$, and $w_3$ will be computed as
$$
    e_1 = \frac{1}{2}\left(e_{11} + e_{12}\right),\;\;\;
    e_2 = \frac{1}{3}\left(e_{21} + e_{22} + e_{23}\right),\;\;\;
    e_3 = e_{31}.
$$
Here, $e_.$ is the final representation of token $t_.$ or word $w_.$. Moreover, for the models that have gone through MNTP, we calculate the representation based on sub-tokens of the previous word. Using the same example as above, for models trained with MNTP, the representation of words $w_1$, $w_2$, and $w_3$ will be computed as:
$$
    e_1 = \frac{1}{2}\left(e_{\text{BOS}} + e_{11}\right),\;\;\;
    e_2 = \frac{1}{3}\left(e_{12} + e_{21} + e_{22}\right),\;\;\;
    e_3 = e_{23}.
$$
\subsection{Additional results}

\subsubsection{Word-level task results}
\label{subsec:word-level-results}

We present the detailed breakdown of the performance of LLM2Vec-transformed models on the word-level tasks in \Cref{tab:appendix:word-task-results}. Our results show that applying MNTP training to decoder-only LLMs helps them take advantage of the enabled bidirectional attention which boosts their performance on word-level tasks.

\begin{table}[h]
    \centering
    \small
    \begin{tabular}{lccc}
    \toprule
    \textbf{Model} & \textbf{Chunking} & \textbf{NER} & \textbf{POS tagging}\\
    \midrule
    \texttt{Encoder-only}\\\cmidrule{1-1}
    BERT-large & 71.77 & 90.09 & 75.12 \\
    DeBERTa-large & 85.74 & 94.97 & 86.49 \\
    \cmidrule{1-1}
    \texttt{\sllama{}}\\\cmidrule{1-1}
    Uni & 86.10 & 96.09 & 90.89 \\
    Bi & 76.50 & 92.17 & 89.18 \\
    Bi + MNTP & \textbf{90.51} & \textbf{96.59} & \textbf{92.04} \\
    Bi + SimCSE  & 75.93 & 91.45 & 89.22 \\
    Bi + MNTP + SimCSE & 89.33 & 95.90 & 90.38 \\
    \cmidrule{1-1}
    \texttt{\llama{}}\\\cmidrule{1-1}
    Uni & 88.23 & 96.59 & 91.53 \\
    Bi & 78.24 & 92.31 & 90.62 \\
    Bi + MNTP & \textbf{91.61} & \textbf{97.16} & \textbf{92.61} \\
    Bi + SimCSE  & 77.75 & 91.96 & 90.48 \\
    Bi + MNTP + SimCSE & 89.66 & 96.05 & 90.53 \\
    \cmidrule{1-1}
    \texttt{\mistral{}}\\\cmidrule{1-1}
    Uni & 87.53 & 96.52 & 90.86 \\
    Bi & 85.66 & 97.14 & 90.70 \\
    Bi + MNTP & \textbf{91.17} & \textbf{97.18} & \textbf{92.35} \\
    Bi + SimCSE & 86.91 & 97.15 & 92.12 \\
    Bi + MNTP + SimCSE & 90.69 & 96.87 & 92.08 \\
    \bottomrule
    \end{tabular}
    \caption{Unsupervised results on the word-level tasks for different models.}
    \label{tab:appendix:word-task-results}
\end{table}

\subsubsection{Sentence-level task results}
\label{subsec:sentence-level-results}

\Cref{tab:appendix:unsupervised-pooling-results} presents the results on MTEB subset for all models across different pooling methods. Results show that While weighted mean works the best for causal (i.e., Uni) models, mean pooling performs the best for LLM2Vec approach.

\begin{table}[ht]
    \centering
    \begin{tabular}{lccc}
    \toprule
    \textbf{Model} & \textbf{EOS} & \textbf{Mean} & \textbf{W. mean}\\
    \midrule
    \texttt{\sllama{}}\\\cmidrule{1-1}
    Uni & 27.72 & 33.03 & 34.99 \\
    Bi & 21.16 & 30.26 & 30.20 \\
    Bi + MNTP & 29.16 & 42.10 & 38.67 \\
    Uni + SimCSE & 37.44 & 44.95 & 47.13 \\
    Bi + SimCSE & 40.43 & 44.46 & 44.83 \\
    Bi + MNTP + SimCSE & 45.57 & \textbf{52.40} & 50.23 \\
    \cmidrule{1-1}
    \texttt{\llama{}}\\\cmidrule{1-1}
    Uni & 33.23 & 45.83 & 47.85 \\
    Bi & 34.47 & 38.22 & 37.50 \\
    Bi + MNTP & 32.66 & 48.00 & 44.30 \\
    Uni + SimCSE & 38.47 & 52.03 & 53.55 \\
    Bi + SimCSE & 40.37 & 44.13 & 44.08 \\
    Bi + MNTP + SimCSE & 50.61 & \textbf{58.97} & 55.75 \\
    \cmidrule{1-1}
    \texttt{\mistral{}}\\\cmidrule{1-1}
    Uni & 22.12 & 43.00 & 44.01 \\
    Bi & 25.17 & 50.07 & 45.20 \\
    Bi + MNTP & 26.54 & 53.89 & 48.93 \\
    Uni + SimCSE & 34.60 & 52.04 & 53.95 \\
    Bi + SimCSE & 49.73 & 60.29 & 56.56 \\
    Bi + MNTP + SimCSE & 53.67 & \textbf{60.50} & 57.55 \\
    \bottomrule
    
    \end{tabular}
    \caption{Unsupervised results on MTEB subset for different models.}
    \label{tab:appendix:unsupervised-pooling-results}
\end{table}

In \Cref{tab:appendix:sentence-task-results}, we additionally present a breakdown of the unsupervised performance of LLM2Vec-transformed models on MTEB.
\section{Comparison with Echo embedding}
\label{sec:appendix:echo}

\subsection{Reproducibility}
\label{subsec:appendix:echo-reproducibility}

Concurrent to our work, \citet{springer2024repetition} proposed Echo embeddings, a simple approach to convert decoder-only LLMs into text embedders by copying the input sequence and appending it to itself. For evaluation, they follow a prompt sampling procedure for the task instruction. However, they report that the exact wording or template used as a prompting strategy does not have a strong effect on the performance.


For a fair comparison to our proposed models, we implement Echo embeddings using the instructions in our evaluation setup (\Cref{subsec:appendix:mteb-instruction}).
To do a sanity check on our implementation, as well as to see the impact of exact wording of instructions, we evaluate Echo embedding on the same subset of 26 MTEB tasks that was chosen in their work. We run this evaluation using the \texttt{Mistral-7B-Instruct-v0.1} model to ensure that the results are directly comparable to theirs. 

The unsupervised Echo model based on our implementation and instructions achieved a score of $55.22$ on the 26 task subset, whereas their reported score is $55.07$. This result validates our implementation and confirms an observation made by \citet{springer2024repetition} -- the exact wording or template used does not have a strong effect on the performance.

\subsection{Efficiency}
\label{subsec:appendix:echo-efficiency}

In \Cref{tab:appendix:echo:time}, we report the approximate evaluation time it took (in hours) to evaluate each of the models on MTEB using 8x 80GB A100 GPUs. Given that Echo embeddings rely on copying the input text, evaluation takes much longer compared to our approach. We note that the increased inference time of Echo embeddings is especially problematic for the encoding of large retrieval corpora in MTEB benchmark.

\begin{table}[h]
    \centering
    \begin{tabular}{lcc}
        \toprule
        \textbf{Model} & \textbf{LLM2Vec} & \textbf{Echo embeddings}\\ \midrule
        \sllama{} & $\approx 30$ hrs & $\approx 40$ hrs\\\midrule
        \llama{} & $\approx 42$ hrs & $\approx 63$ hrs\\\midrule
        \mistral{} & $\approx 44$ hrs & $\approx 64$ hrs\\
        \bottomrule
    \end{tabular}
    \caption{Evaluation time of Echo Embeddings compared to LLM2Vec in hours on 8x 80GB A100 GPUs.}
    \label{tab:appendix:echo:time}
\end{table}

\section{More analysis results} 
\label{sec:appendix:analysis}

\paragraph{Data used for our analysis}

\Cref{tab:appendix:echo-analysis-examples} shows examples of the data used for our cosine similarity analysis in \Cref{sec:llm2vec-analysis}.

\begin{table}[ht]
    \centering
    \begin{tabular}{p{4.2cm}p{4.2cm}p{4.2cm}}
    \toprule
        $q$: \textbf{the query} & $s^+$: \textbf{the positive sample} & $s^-$: \textbf{the negative sample} \\ \midrule
        \textit{She loves to travel in summer,} especially to cold destinations, avoiding hot and crowded places. & \textit{She loves to travel in summer,} specifically to chilly locations, steering clear of warm, populous areas. & \textit{She loves to travel in summer,} but prefers to visit hot and bustling tourist spots.\\ \midrule
        \textit{The cat often sits by the window,} dreaming of chasing birds and enjoying the warm sunshine. & \textit{The cat often sits by the window,} imagining bird pursuits and basking in the sunlight. & \textit{The cat often sits by the window,} but is too lazy to dream of chasing anything.\\ \midrule
        \textit{He reads books every night,} finding solace in fiction and escaping from the stresses of daily life. & \textit{He reads books every night,} seeking comfort in stories and evading everyday tensions. & \textit{He reads books every night,} yet he feels that non-fiction is more engaging and informative.\\ \midrule
        \textit{She paints landscapes on weekends,} expressing her creativity through vibrant colors and abstract forms. & \textit{She paints landscapes on weekends,} showcasing her artistic flair with lively hues and unconventional shapes. & \textit{She paints landscapes on weekends,} preferring realistic and detailed depictions of nature.\\ \bottomrule
    \end{tabular}
    \caption{Toy data used for the analysis in \Cref{sec:llm2vec-analysis}. These sentences were originally collected by \citet{springer2024repetition}.}
    \label{tab:appendix:echo-analysis-examples}
\end{table}

\paragraph{Cosine similarity analysis}

\begin{figure}[h]
    \centering
    \begin{subfigure}[t]{1.0\textwidth}
        \centering
        \begin{subfigure}[t]{0.24\textwidth}
            \centering
            \includegraphics[width=0.99\textwidth]{Plots/embedding_analysis/Sheared-LLaMA-1.3B_bi_initialize_skip-True.pdf}
        \end{subfigure}%
        \begin{subfigure}[t]{0.24\textwidth}
            \centering
            \includegraphics[width=0.99\textwidth]{Plots/embedding_analysis/Sheared-LLaMA-1.3B_mlm_initialize_skip-True.pdf}
        \end{subfigure}
        \begin{subfigure}[t]{0.24\textwidth}
            \centering
            \includegraphics[width=0.99\textwidth]{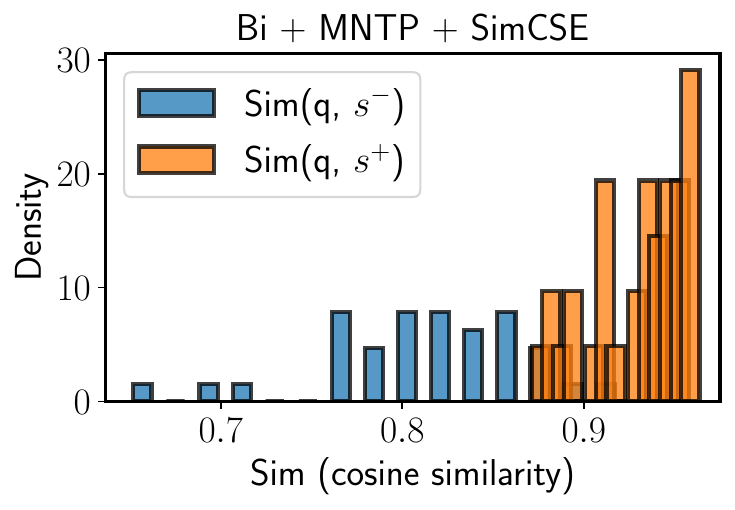}
        \end{subfigure}
        \begin{subfigure}[t]{0.24\textwidth}
            \centering
            \includegraphics[width=0.99\textwidth]{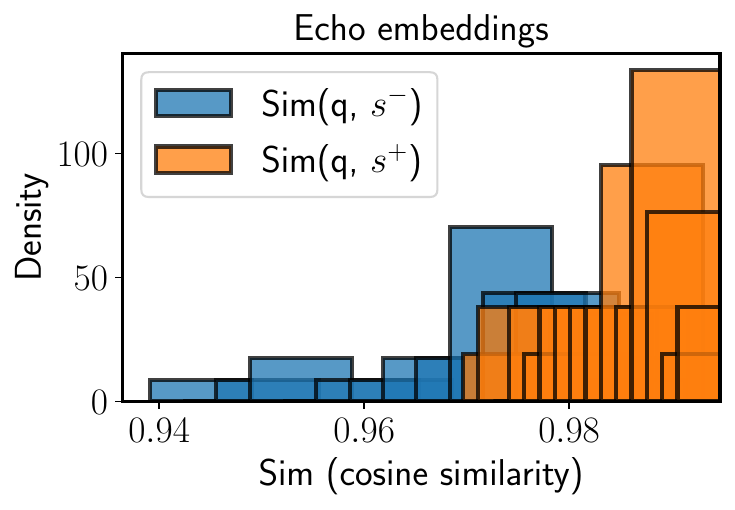}
        \end{subfigure}%
        \caption{\texttt{S-LLaMA-1.3B}}
    \end{subfigure}
    \\
    \begin{subfigure}[t]{1.0\textwidth}
        \begin{subfigure}[t]{0.24\textwidth}
            \centering
            \includegraphics[width=0.99\textwidth]{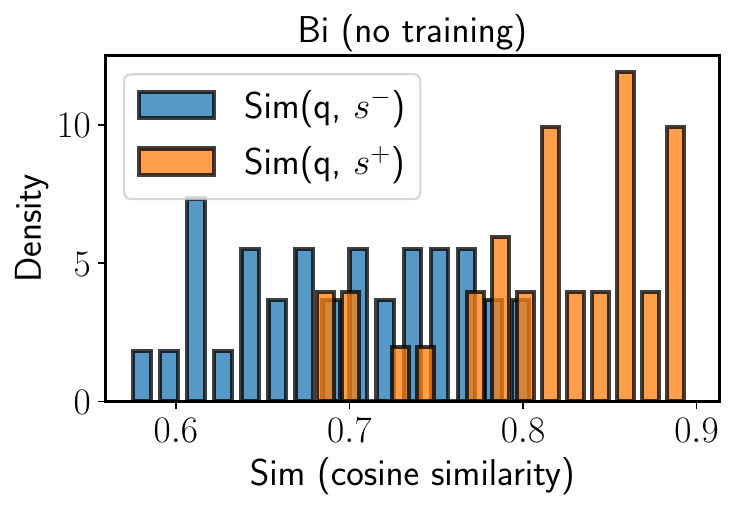}
        \end{subfigure}%
        \begin{subfigure}[t]{0.24\textwidth}
            \centering
            \includegraphics[width=0.99\textwidth]{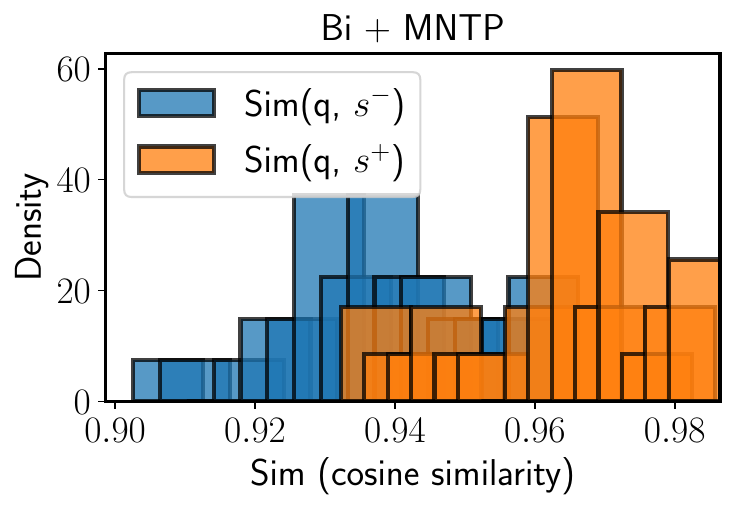}
        \end{subfigure} 
        \begin{subfigure}[t]{0.24\textwidth}
            \centering
            \includegraphics[width=0.99\textwidth]{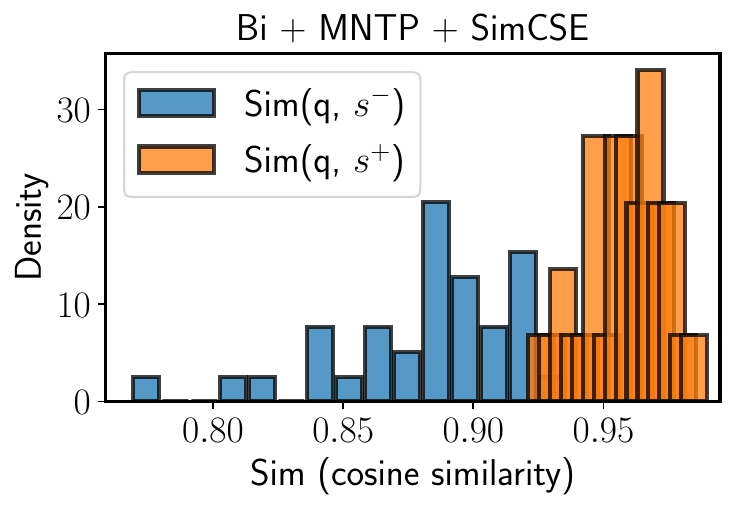}
        \end{subfigure}%
        \begin{subfigure}[t]{0.24\textwidth}
            \centering
            \includegraphics[width=0.99\textwidth]{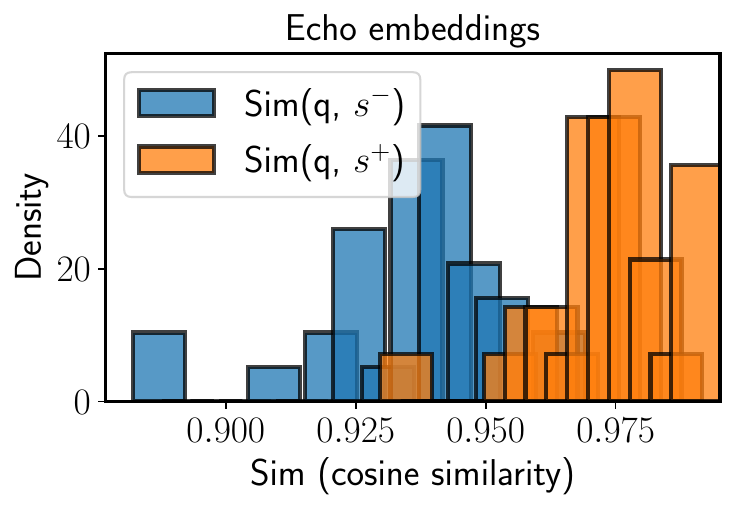}
        \end{subfigure} 
        \caption{\texttt{LLaMA-2-7B}}
    \end{subfigure}
    \\
    \begin{subfigure}[t]{1.0\textwidth}
        \begin{subfigure}[t]{0.24\textwidth}
            \centering
            \includegraphics[width=0.99\textwidth]{Plots/embedding_analysis/Mistral-7B-Instruct-v0.2_bi_initialize_skip-True.pdf}
        \end{subfigure}%
        \begin{subfigure}[t]{0.24\textwidth}
            \centering
            \includegraphics[width=0.99\textwidth]{Plots/embedding_analysis/Mistral-7B-Instruct-v0.2_mlm_initialize_skip-True.pdf}
        \end{subfigure} 
        \begin{subfigure}[t]{0.24\textwidth}
            \centering
            \includegraphics[width=0.99\textwidth]{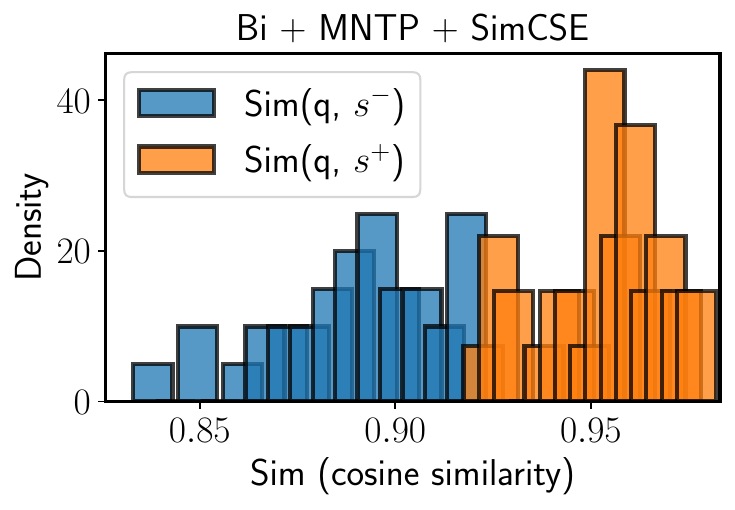}
        \end{subfigure}%
        \begin{subfigure}[t]{0.24\textwidth}
            \centering
            \includegraphics[width=0.99\textwidth]{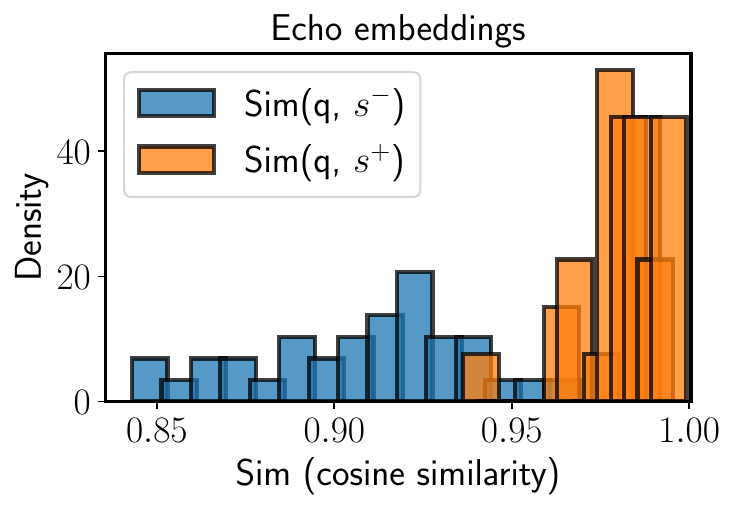}
        \end{subfigure} 
        \caption{\texttt{Mistral-7B}}
    \end{subfigure}
    
    \caption{Cosine similarity between query and negative as well as positive examples for \sllama{}, \llama{}, and \mistral{}.  
    }
    \label{fig:appendix:cosine-analysis}
\end{figure}

\Cref{fig:appendix:cosine-analysis} provide cosine similarity results for all three models. In addition to our LLM2Vec-transformed models, we also provide results for Echo emebddings.

\subsection{Additional plots}

\begin{figure}[h]
    \centering
    \begin{subfigure}[t]{0.33\textwidth}
        \centering
        \includegraphics[width=0.90\textwidth]{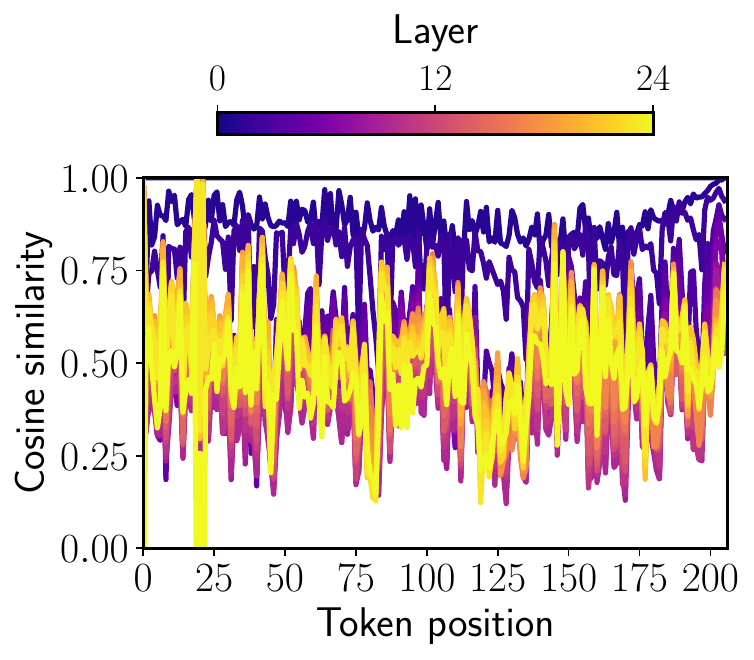}
        \caption{\texttt{S-LLaMA-1.3B}}
    \end{subfigure}%
    ~
    \begin{subfigure}[t]{0.33\textwidth}
        \centering
        \includegraphics[width=0.90\textwidth]{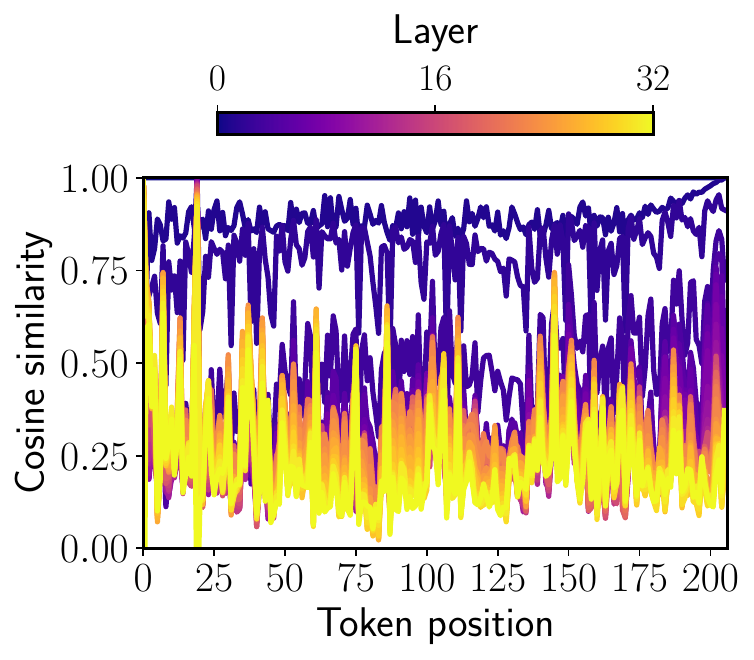}
        \caption{\texttt{Llama-2-7B}}
    \end{subfigure}%
    ~
    \begin{subfigure}[t]{0.33\textwidth}
        \centering
        \includegraphics[width=0.90\textwidth]{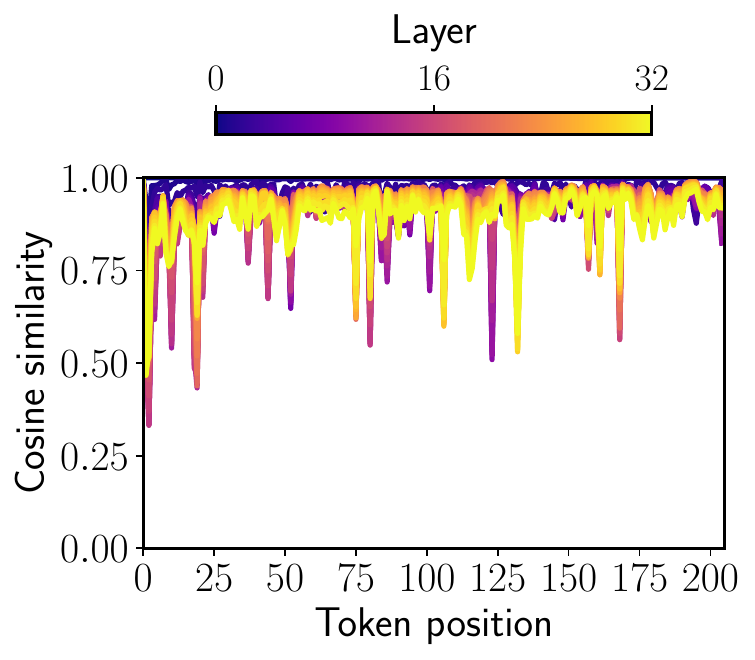}
        \caption{\texttt{Mistral-7B}}
    \end{subfigure}  
    \\
    \begin{subfigure}[t]{0.33\textwidth}
        \centering
        \includegraphics[width=0.90\textwidth]{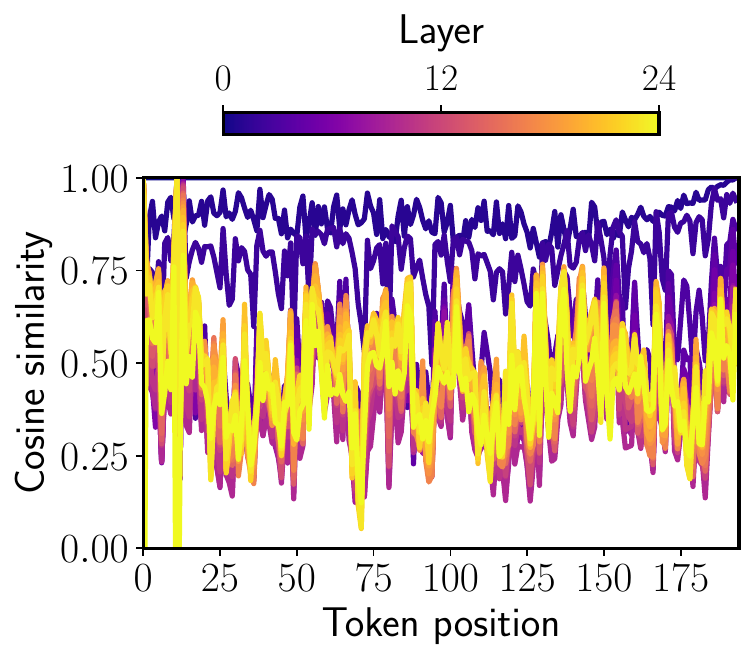}
        \caption{\texttt{S-LLaMA-1.3B}}
    \end{subfigure}%
    ~
    \begin{subfigure}[t]{0.33\textwidth}
        \centering
        \includegraphics[width=0.90\textwidth]{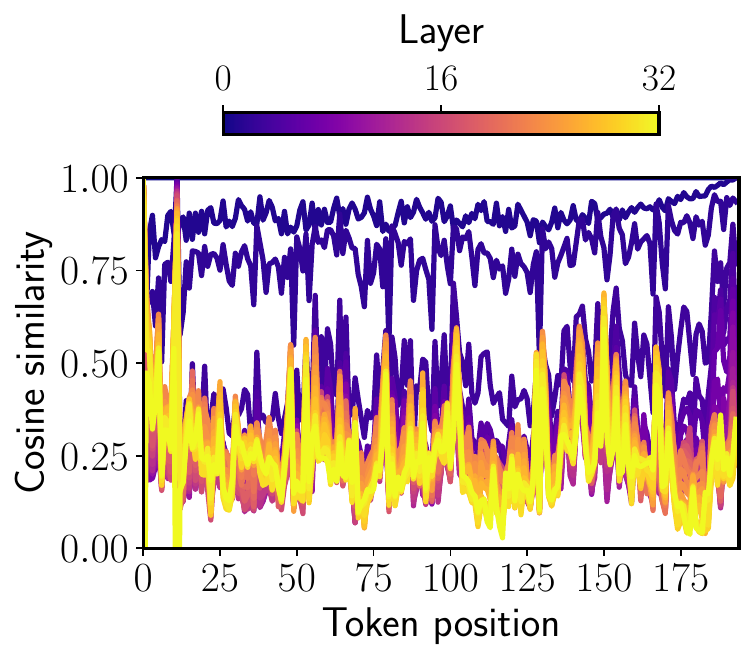}
        \caption{\texttt{Llama-2-7B}}
    \end{subfigure}%
    ~
    \begin{subfigure}[t]{0.33\textwidth}
        \centering
        \includegraphics[width=0.90\textwidth]{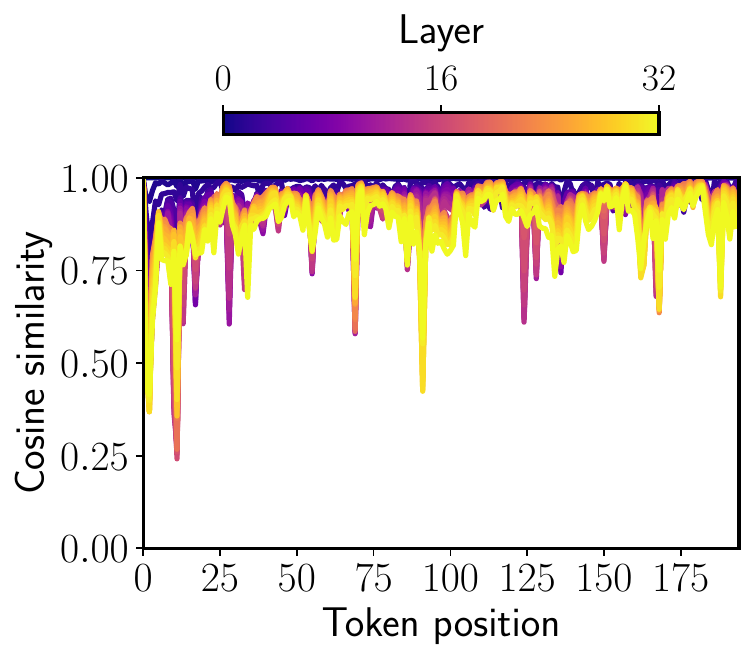}
        \caption{\texttt{Mistral-7B}}
    \end{subfigure}  
    \caption{Cosine similarities at different token positions at layers when comparing representations constructed with causal attention to those constructed with bidirectional attention (without training).}
    \label{fig:appendix:hidden_analysis}
\end{figure}

\begin{figure}[h]
    \centering
    \begin{subfigure}[t]{0.33\textwidth}
        \centering
        \includegraphics[width=0.90\textwidth]{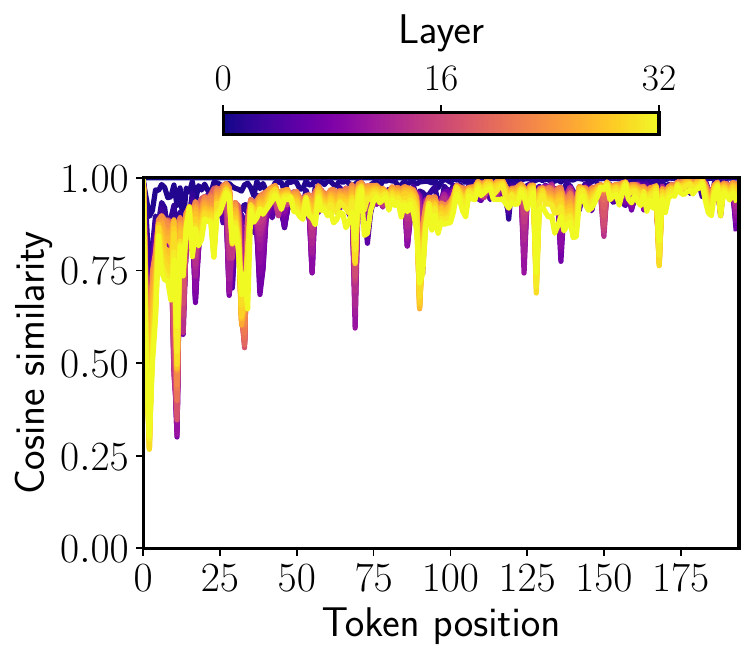}
        \caption{\texttt{Mistral-7B-v0.1}}
    \end{subfigure}%
    ~
    \begin{subfigure}[t]{0.33\textwidth}
        \centering
        \includegraphics[width=0.90\textwidth]{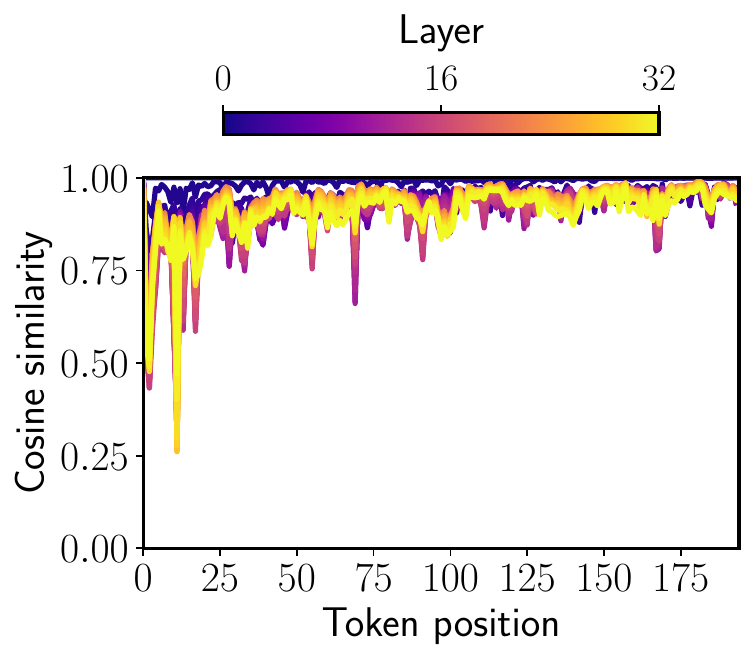}
        \caption{\texttt{Mistral-7B-Instruct-v0.1}}
    \end{subfigure}%
    ~
    \begin{subfigure}[t]{0.33\textwidth}
        \centering
        \includegraphics[width=0.90\textwidth]{Plots/hidden_analysis/Baltimore_Mistral-7B-Instruct-v0.2.pdf}
        \caption{\texttt{Mistral-7B-Instruct-v0.2}}
    \end{subfigure}  
    \caption{Cosine similarities at different token positions at layers when comparing representations of Mistral models constructed with causal attention to those constructed with bidirectional attention (without training).}
    \label{fig:appendix:hidden_analysis_mistral}
\end{figure}

\paragraph{Representation analysis}

\Cref{fig:appendix:hidden_analysis} provides additional plots for the representation analysis using two different Wikipedia paragraphs. The trends closely follow those reported in \Cref{sec:llm2vec-analysis}. \Cref{fig:appendix:hidden_analysis_mistral} shows that the same behavior we observe for \texttt{Mistral-7B-instruct-v0.2} also holds true for other variants of the Mistral-7B model. We take this as additional evidence that the Mistral-7B base model was trained with some for of bidirectional attention.

\section{Details on supervised results}
\label{sec:appendix:supervised-ablations}

\subsection{E5 dataset}
\label{sec:appendix:e5}

The dataset consists of ELI5 (sample ratio 0.1) \citep{fan-etal-2019-eli5}, HotpotQA \citep{yang-etal-2018-hotpotqa}, FEVER \citep{thorne-etal-2018-fever}, MIRACL \citep{zhang-et-all-2023-MIRACL}, MS-MARCO passage ranking (sample ratio 0.5) and document ranking (sample ratio 0.2) \citep{}, NQ \citep{karpukhin-etal-2020-DPR}, NLI \citep{gao-etal-2021-simcse}, SQuAD \citep{rajpurkar-etal-2016-squad},
TriviaQA \citep{joshi-etal-2017-triviaqa},
Quora Duplicate Questions (sample ratio 0.1) \citep{quora-question-pairs},
Mr- TyDi \citep{zhang-etal-2021-mr}, DuReader \citep{he-etal-2018-dureader}, and T2Ranking (sample ratio 0.5) \citep{t2ranking}.
The instruction used for each dataset can be found in \Cref{tab:appendix:finetuning_instructions}.
\begin{table}
\scriptsize
\centering
\begin{tabular}{p{0.23\linewidth}p{0.72\linewidth}}
\toprule
\textbf{Dataset} & \textbf{Instruction(s)} \\
\midrule
NLI & Given a premise, retrieve a hypothesis that is entailed by the premise \\
    & Retrieve semantically similar text \\
DuReader & Given a Chinese search query, retrieve web passages that answer the question \\
ELI5 & Provided a user question, retrieve the highest voted answers on Reddit ELI5 forum \\
FEVER & Given a claim, retrieve documents that support or refute the claim \\
HotpotQA & Given a multi-hop question, retrieve documents that can help answer the question \\
MIRACL & Given a question, retrieve Wikipedia passages that answer the question \\
MrTyDi & Given a question, retrieve Wikipedia passages that answer the question \\
MSMARCO Passage & Given a web search query, retrieve relevant passages that answer the query \\
MSMARCO Document & Given a web search query, retrieve relevant documents that answer the query \\
NQ & Given a question, retrieve Wikipedia passages that answer the question \\
QuoraDuplicates & Given a question, retrieve questions that are semantically equivalent to the given question \\
 & Find questions that have the same meaning as the input question \\
SQuAD & Retrieve Wikipedia passages that answer the question \\
T2Ranking & Given a Chinese search query, retrieve web passages that answer the question \\
TriviaQA & Retrieve Wikipedia passages that answer the question \\
\bottomrule
\end{tabular}
\caption{Instructions used for each of the E5 datasets.}
\label{tab:appendix:finetuning_instructions}
\end{table}

\subsection{Training details}
\label{subsec:sup-training-details}

All models are trained with LoRA $r = 16$ and $\alpha = 32$, brain floating point (bfloat16) quantization, gradient checkpointing, and flash attention 2 \citep{dao2023flashattention2} to optimize GPU memory consumption. We train on 8 NVIDIA A100 GPUs with an effective batch size of 512 for 1000 steps using a maximum sequence length of 512 tokens. We use the Adam optimizer with a learning rate of $2e-4$ and a linear learning rate warm-up for the first 300 steps.

\subsection{Results}
\label{sec:appendix:sup-mteb-results}

\Cref{tab:supervised} presents the performance of applying \texttt{Bi + MNTP} and \texttt{Bi + MNTP + SimCSE} with mean pooling on MTEB benchmark. We also compare the performance of our models with recent and popular models trained with only publicly available data. We further report the current top-10 models in the MTEB leaderboard, including LLM2Vec$_\text{\mistral{}}$ in \Cref{tab:supervised-top-10}. Our models achieve the 6th score in the MTEB leaderboard and the 1st among the models trained with only public data. 

\begin{table*}[t]
    \centering
    \small
    \resizebox{\textwidth}{!}{
    \begin{tabular}{cl|ccc|cccccccc}
    \toprule
    \multicolumn{1}{l}{\multirow{2}{*}{\textbf{Rank}}} & \multicolumn{1}{l|}{\multirow{2}{*}{\textbf{Model}}} & \textbf{Size} & \textbf{Public} & \textbf{Embed.} & 
    \textbf{Retr}. & \textbf{Rerank.} & \textbf{Clust.} & \textbf{PairClass.} & \textbf{Class.} & \textbf{STS} & \textbf{Summ.} & \textbf{Avg} \\
    && \multicolumn{1}{c}{\textbf{(GB)}} & \multicolumn{1}{c}{\textbf{Data}} & \multicolumn{1}{c|}{\textbf{Dim.}} & \multicolumn{1}{c}{15}     & \multicolumn{1}{c}{4}     & \multicolumn{1}{c}{11}   & \multicolumn{1}{c}{3}      & \multicolumn{1}{c}{12}    & \multicolumn{1}{c}{10}  & \multicolumn{1}{c}{1}    & \multicolumn{1}{c}{56}  \\ 
    \midrule
    1 & SFR-Embedding-Mistral & 14.22 & $\times$ & 4096 & 59.00 & 60.64 & 51.67 & 88.54 & 78.33 & 85.05 & 31.16 & 67.56 \\
    2 & voyage-lite-02-instruct & 2.45 & $\times$ & 1024 & 56.60 & 58.24 & 52.42 & 86.87 & 79.25 & 85.79 & 31.01 & 67.13 \\
    3 & GritLM-7B & 14.48 & $\times$ & 4096 & 57.41 & 60.49 & 50.61 & 87.16 & 79.46 & 83.35 & 30.37 & 66.76 \\
    4 & e5-mistral-7b-instruct & 14.22 & $\times$ & 4096 & 56.89 & 60.21 & 50.26 & 88.34 & 78.47 & 84.63 & 31.40 & 66.63 \\
    5 & GritLM-8x7B & 93.41 & $\times$ & 4096 & 55.09 & 59.80 & 50.14 & 84.97 & 78.53 & 83.26 & 29.82 & 65.66 \\
    \rowcolor{green!20}
    6 & Bi + MNTP + Mean & 14.22 & $\checkmark$ & 4096 & 55.99 & 58.42 & 45.54 & 87.99 & 76.63 & 84.09 & 29.96 & 64.80\\
    \rowcolor{green!20}
    6 & Bi + MNTP + SimCSE + Mean & 14.22 & $\checkmark$ & 4096 & 56.05 & 58.59 & 45.12 & 88.18 & 76.72 & 83.69 & 30.66 & 64.72 \\
    7 & echo-mistral-7b-instruct-lasttoken & 14.22 & $\checkmark$ & 4096 & 55.52 & 58.14 & 46.32 & 87.34 & 77.43 & 82.56 & 30.73 & 64.68 \\
    8 & mxbai-embed-large-v1 & 0.67 & $\times$ & 1024 & 54.39 & 60.11 & 46.71 & 87.20 & 75.64 & 85.00 & 32.71 & 64.68 \\
    9 & UAE-Large-V1 & 1.34 & $\times$ & 1024 & 54.66 & 59.88 & 46.73 & 87.25 & 75.58 & 84.54 & 32.03 & 64.64 \\
    10 & text-embedding-3-large & - & $\times$ & 3072 & 55.44 & 59.16 & 49.01 & 85.72 & 75.45 & 81.73 & 29.92 & 64.59 \\    
    \bottomrule
     
    \end{tabular}%
    }
    \caption{Top-10 models on the MTEB leaderboard as of 2024-03-29. LLM2Vec achieves the 6th rank overall, and the top rank among models trained with only publicly available data.
    }
    \label{tab:supervised-top-10}
\end{table*}

\begin{table}
\centering   
\scriptsize
\begin{tabular}{p{0.24\linewidth}p{0.7\linewidth}}
\toprule
\textbf{Task Name} & \textbf{Instruction} \\
\midrule
AmazonCounterfactualClassif. & Classify a given Amazon customer review text as either counterfactual or not-counterfactual \\
AmazonPolarityClassification & Classify Amazon reviews into positive or negative sentiment  \\
AmazonReviewsClassification & Classify the given Amazon review into its appropriate rating category  \\
Banking77Classification & Given a online banking query, find the corresponding intents  \\
EmotionClassification &  Classify the emotion expressed in the given Twitter message into one of the six emotions: anger, fear, joy, love, sadness, and surprise  \\
ImdbClassification & Classify the sentiment expressed in the given movie review text from the IMDB dataset  \\
MassiveIntentClassification & Given a user utterance as query, find the user intents  \\
MassiveScenarioClassification & Given a user utterance as query, find the user scenarios  \\
MTOPDomainClassification & Classify the intent domain of the given utterance in task-oriented conversation  \\
MTOPIntentClassification & Classify the intent of the given utterance in task-oriented conversation  \\
ToxicConversationsClassif. & Classify the given comments as either toxic or not toxic  \\
TweetSentimentClassification & Classify the sentiment of a given tweet as either positive, negative, or neutral  \\
ArxivClusteringP2P & Identify the main and secondary category of Arxiv papers based on the titles and abstracts  \\
ArxivClusteringS2S & Identify the main and secondary category of Arxiv papers based on the titles  \\
BiorxivClusteringP2P & Identify the main category of Biorxiv papers based on the titles and abstracts  \\
BiorxivClusteringS2S & Identify the main category of Biorxiv papers based on the titles  \\
MedrxivClusteringP2P & Identify the main category of Medrxiv papers based on the titles and abstracts  \\
MedrxivClusteringS2S & Identify the main category of Medrxiv papers based on the titles  \\
RedditClustering & Identify the topic or theme of Reddit posts based on the titles  \\
RedditClusteringP2P & Identify the topic or theme of Reddit posts based on the titles and posts  \\
StackExchangeClustering & Identify the topic or theme of StackExchange posts based on the titles  \\
StackExchangeClusteringP2P & Identify the topic or theme of StackExchange posts based on the given paragraphs  \\
TwentyNewsgroupsClustering & Identify the topic or theme of the given news articles  \\
SprintDuplicateQuestions & Retrieve duplicate questions from Sprint forum  \\
TwitterSemEval2015 & Retrieve tweets that are semantically similar to the given tweet  \\
TwitterURLCorpus & Retrieve tweets that are semantically similar to the given tweet  \\
AskUbuntuDupQuestions & Retrieve duplicate questions from AskUbuntu forum  \\
MindSmallReranking & Retrieve relevant news articles based on user browsing history  \\
SciDocsRR & Given a title of a scientific paper, retrieve the titles of other relevant papers  \\
StackOverflowDupQuestions & Retrieve duplicate questions from StackOverflow forum  \\
ArguAna & Given a claim, find documents that refute the claim  \\
ClimateFEVER & Given a claim about climate change, retrieve documents that support or refute the claim  \\
CQADupstackRetrieval &  Given a question, retrieve detailed question descriptions from Stackexchange that are duplicates to the given question  \\
DBPedia & Given a query, retrieve relevant entity descriptions from DBPedia  \\
FEVER & Given a claim, retrieve documents that support or refute the claim  \\
FiQA2018 & Given a financial question, retrieve user replies that best answer the question  \\
HotpotQA & Given a multi-hop question, retrieve documents that can help answer the question  \\
MSMARCO & Given a web search query, retrieve relevant passages that answer the query  \\
NFCorpus & Given a question, retrieve relevant documents that best answer the question  \\
NQ & Given a question, retrieve Wikipedia passages that answer the question  \\
QuoraRetrieval & Given a question, retrieve questions that are semantically equivalent to the given question  \\
SCIDOCS & Given a scientific paper title, retrieve paper abstracts that are cited by the given paper  \\
SciFact & Given a scientific claim, retrieve documents that support or refute the claim  \\
Touche2020 & Given a question, retrieve detailed and persuasive arguments that answer the question  \\
TRECCOVID & Given a query on COVID-19, retrieve documents that answer the query  \\
STS* & Retrieve semantically similar text.  \\
BUCC/Tatoeba & Retrieve parallel sentences.  \\
SummEval & Given a news summary, retrieve other semantically similar summaries  \\
\bottomrule
\end{tabular}
\caption{Instructions used for evaluation on the MTEB benchmark.
``STS*'' refers to all the STS tasks.} \label{tab:appendix:mteb_instructions}
\end{table}
\begin{table}[t]
    \centering
    \small
    \begin{tabular}{l|ccc}
    \toprule
    \textbf{Task} & \sllama{} & \llama{} & \mistral{} \\
    \midrule
    AmazonCounterfactualClassification & 72.93 & 76.91 & 76.94 \\
    AmazonPolarityClassification & 74.28 & 79.05 & 85.29 \\
    AmazonReviewsClassification & 36.14 & 40.08 & 47.09 \\
    ArguAna & 43.64 & 47.09 & 51.00 \\
    ArxivClusteringP2P & 42.91 & 47.81 & 47.56 \\
    ArxivClusteringS2S & 35.20 & 40.53 & 39.92 \\
    AskUbuntuDupQuestions & 52.70 & 55.56 & 58.60 \\
    BIOSSES & 75.12 & 82.41 & 83.29 \\
    Banking77Classification & 79.00 & 84.65 & 86.16 \\
    BiorxivClusteringP2P & 35.02 & 38.12 & 36.14 \\
    BiorxivClusteringS2S & 27.21 & 31.25 & 30.26 \\
    CQADupstackRetrieval & 18.50 & 30.78 & 33.37 \\
    ClimateFEVER & 18.95 & 20.67 & 22.97 \\
    DBPedia & 13.21 & 25.81 & 25.48 \\
    EmotionClassification & 42.85 & 46.58 & 48.88 \\
    FEVER & 16.96 & 43.48 & 45.11 \\
    FiQA2018 & 16.99 & 24.62 & 27.24 \\
    HotpotQA & 22.64 & 48.46 & 54.54 \\
    ImdbClassification & 71.92 & 75.68 & 77.95 \\
    MSMARCO & \phantom{0}7.03 & 18.81 & 19.13 \\
    MTOPDomainClassification & 91.24 & 94.33 & 95.48 \\
    MTOPIntentClassification & 74.08 & 79.54 & 82.84 \\
    MassiveIntentClassification & 69.99 & 73.84 & 76.65 \\
    MassiveScenarioClassification & 75.15 & 79.17 & 79.99 \\
    MedrxivClusteringP2P & 30.15 & 30.94 & 30.11 \\
    MedrxivClusteringS2S & 26.96 & 28.04 & 26.93 \\
    MindSmallReranking & 29.52 & 30.86 & 29.73 \\
    NFCorpus & 15.73 & 26.81 & 27.16 \\
    NQ & 17.96 & 33.21 & 34.16 \\
    QuoraRetrieval & 78.23 & 86.15 & 84.40 \\
    RedditClustering & 38.67 & 42.84 & 41.83 \\
    RedditClusteringP2P & 53.42 & 60.10 & 62.08 \\
    SCIDOCS & \phantom{0}5.53 & 10.00 & 15.35 \\
    SICK-R & 69.34 & 71.77 & 75.55 \\
    STS12 & 60.09 & 65.39 & 67.65 \\
    STS13 & 72.52 & 79.26 & 83.90 \\
    STS14 & 66.70 & 72.98 & 76.97 \\
    STS15 & 77.69 & 82.72 & 83.80 \\
    STS16 & 75.94 & 81.02 & 81.91 \\
    STS17 & 81.67 & 86.70 & 85.58 \\
    STS22 & 63.70 & 63.47 & 65.93 \\
    STSBenchmark & 73.36 & 78.32 & 80.42 \\
    SciDocsRR & 67.76 & 77.62 & 77.81 \\
    SciFact & 38.31 & 64.48 & 68.67 \\
    SprintDuplicateQuestions & 77.36 & 87.57 & 91.30 \\
    StackExchangeClustering & 59.35 & 65.12 & 67.34 \\
    StackExchangeClusteringP2P & 31.47 & 33.61 & 34.50 \\
    StackOverflowDupQuestions & 40.82 & 47.77 & 49.80 \\
    SummEval & 31.23 & 31.38 & 30.19 \\
    TRECCOVID & 56.04 & 60.67 & 55.66 \\
    Touche2020 & 19.17 & 10.18 & \phantom{0}6.54 \\
    ToxicConversationsClassification & 68.41 & 71.81 & 70.71 \\
    TweetSentimentExtractionClassification & 56.08 & 57.17 & 60.90 \\
    TwentyNewsgroupsClustering & 31.54 & 30.76 & 30.26 \\
    TwitterSemEval2015 & 61.54 & 65.14 & 68.76 \\
    TwitterURLCorpus & 77.73 & 80.94 & 82.76 \\
    \midrule
    Average & 49.42 & 55.36 & 56.80 \\
    \bottomrule
    \end{tabular}
    \caption{Unsupervised results of LLM2Vec transformed models on MTEB.}\label{tab:appendix:sentence-task-results}
\end{table}
\begin{table}[t]
    \centering
    \small
    \begin{tabular}{l|ccc}
    \toprule
    \textbf{Task} & \sllama{} & \llama{} & \mistral{} \\
    \midrule
    AmazonCounterfactualClassification & 77.42 & 82.22 & 77.58 \\
    AmazonPolarityClassification & 82.05 & 89.69 & 91.12 \\
    AmazonReviewsClassification & 40.81 & 48.47 & 49.97 \\
    ArguAna & 51.66 & 56.53 & 57.48 \\
    ArxivClusteringP2P & 43.47 & 43.14 & 42.81 \\
    ArxivClusteringS2S & 39.85 & 42.38 & 44.24 \\
    AskUbuntuDupQuestions & 60.71 & 63.13 & 63.98 \\
    BIOSSES & 85.88 & 82.13 & 85.24 \\
    Banking77Classification & 86.01 & 88.17 & 88.31 \\
    BiorxivClusteringP2P & 37.10 & 35.88 & 34.27 \\
    BiorxivClusteringS2S & 34.28 & 34.81 & 35.53 \\
    CQADupstackRetrieval & 41.73 & 45.94 & 48.84 \\
    ClimateFEVER & 33.49 & 30.70 & 35.19 \\
    DBPedia & 43.58 & 48.42 & 49.58 \\
    EmotionClassification & 48.38 & 51.71 & 52.05 \\
    FEVER & 86.81 & 89.93 & 89.40 \\
    FiQA2018 & 41.00 & 51.28 & 53.11 \\
    HotpotQA & 63.85 & 72.99 & 74.07 \\
    ImdbClassification & 75.33 & 85.78 & 87.42 \\
    MSMARCO & 38.32 & 41.45 & 42.17 \\
    MTOPDomainClassification & 94.09 & 95.57 & 96.04 \\
    MTOPIntentClassification & 77.05 & 82.81 & 84.77 \\
    MassiveIntentClassification & 75.58 & 78.06 & 79.29 \\
    MassiveScenarioClassification & 79.16 & 81.35 & 81.64 \\
    MedrxivClusteringP2P & 33.55 & 32.23 & 31.07 \\
    MedrxivClusteringS2S & 31.11 & 31.37 & 31.27 \\
    MindSmallReranking & 31.96 & 31.34 & 31.50 \\
    NFCorpus & 37.12 & 40.33 & 39.33 \\
    NQ & 53.89 & 61.24 & 61.70 \\
    QuoraRetrieval & 87.37 & 85.59 & 87.75 \\
    RedditClustering & 53.02 & 61.10 & 60.24 \\
    RedditClusteringP2P & 60.47 & 64.52 & 64.12 \\
    SCIDOCS & 17.96 & 21.05 & 22.50 \\
    SICK-R & 82.25 & 83.01 & 83.70 \\
    STS12 & 78.28 & 78.85 & 78.80 \\
    STS13 & 85.52 & 86.84 & 86.37 \\
    STS14 & 82.49 & 84.04 & 84.04 \\
    STS15 & 88.76 & 88.72 & 88.99 \\
    STS16 & 87.11 & 86.79 & 87.22 \\
    STS17 & 90.10 & 90.63 & 90.19 \\
    STS22 & 68.25 & 67.55 & 67.68 \\
    STSBenchmark & 87.16 & 88.72 & 88.65 \\
    SciDocsRR & 79.23 & 84.03 & 83.80 \\
    SciFact & 72.08 & 77.30 & 78.86 \\
    SprintDuplicateQuestions & 96.25 & 96.83 & 96.82 \\
    StackExchangeClustering & 63.04 & 67.98 & 70.73 \\
    StackExchangeClusteringP2P & 34.01 & 33.20 & 34.50 \\
    StackOverflowDupQuestions & 49.61 & 51.02 & 54.41 \\
    SummEval & 30.01 & 28.49 & 29.96 \\
    TRECCOVID & 80.41 & 79.25 & 77.69 \\
    Touche2020 & 22.31 & 16.92 & 22.18 \\
    ToxicConversationsClassification & 69.92 & 71.01 & 69.26 \\
    TweetSentimentExtractionClassification & 60.76 & 61.11 & 62.14 \\
    TwentyNewsgroupsClustering & 49.37 & 51.04 & 52.18 \\
    TwitterSemEval2015 & 76.14 & 80.70 & 80.60 \\
    TwitterURLCorpus & 86.23 & 86.56 & 86.56 \\
    \midrule
    Average & 61.85 & 64.14 & 64.80 \\
    \bottomrule
    \end{tabular}
    \caption{Supervised results of LLM2Vec (only \texttt{Bi + MNTP}) models on MTEB.}\label{tab:appendix:sup-mteb-results}
\end{table}

We present the detailed performance of supervised LLM2Vec-transformed models on full MTEB in \Cref{tab:appendix:sup-mteb-results}. Here, we only report the \texttt{Bi + MNTP} transformed models as we showed they perform the best after supervised fine-tuning.

\end{document}